\documentclass[lettersize,journal]{IEEEtran}

\usepackage{setspace}
\usepackage{indentfirst}
\usepackage[numbers,sort&compress]{natbib}
\usepackage{xcolor}
\usepackage[pdftex]{graphicx}
\usepackage{amsmath}
\usepackage{amsthm}
\usepackage{amssymb}
\usepackage[ruled,linesnumbered]{algorithm2e}
\usepackage{array}
\usepackage[caption=true,font=footnotesize]{subfig}
\usepackage{fixltx2e}
\usepackage{stfloats}
\usepackage{url}
\usepackage{comment}
\usepackage{multirow}
\usepackage{ulem}
\usepackage{verbatim}
\allowdisplaybreaks[2]

\newtheorem{thm}{Theorem}
\newtheorem{lem}{Lemma}

\newtheorem{assum}{Assumption}
\newtheorem{cor}{Corollary}
\theoremstyle{definition}

\usepackage[T1]{fontenc}

\begin{document}
\title{Hierarchical Personalized Federated Learning Over Massive Mobile Edge Computing Networks}
\author{Chaoqun You,~\IEEEmembership{Member,~IEEE,}
        Kun Guo,~\IEEEmembership{Member,~IEEE,}\\
        Howard H. Yang~\IEEEmembership{Member,~IEEE,}
        and~Tony~Q.~S.~Quek,~\IEEEmembership{Fellow,~IEEE}
\thanks{C. You and T. Q. S. Quek are with the Wireless Networks and Design Systems Group, Singapore University of Design and Technology, 487372, Singapore (e-mail: chaoqun\_you, tonyquek@sutd.edu.sg).}
\thanks{K. Guo is with the Shanghai Key Laboratory of Multidimensional Information Processing, School of Communications and Electronics Engineering, East China Normal University, Shanghai 200241, China (e-mail: kguo@cee.ecnu.edu.cn).}
\thanks{H. H. Yang is with the Zhejiang University/University of Illinois at Urbana-Champaign Institute, Zhejiang University, Haining 314400, China (email: haoyang@intl.zju.edu.cn).}}

\markboth{Journal of \LaTeX\ Class Files,~Vol.~14, No.~8, August~2015}%
{Shell \MakeLowercase{\textit{et al.}}: Bare Demo of IEEEtran.cls for IEEE Journals}

\maketitle
\normalem
\begin{abstract}
Personalized Federated Learning (PFL) is a new Federated Learning (FL) paradigm, particularly tackling the heterogeneity issues brought by various mobile user equipments (UEs) in mobile edge computing (MEC) networks. However, due to the ever-increasing number of UEs and the complicated administrative work it brings, it is desirable to switch the PFL algorithm from its conventional two-layer framework to a multiple-layer one. In this paper, we propose hierarchical PFL (HPFL), an algorithm for deploying PFL over massive MEC networks. The UEs in HPFL are divided into multiple clusters, and the UEs in each cluster forward their local updates to the edge server (ES) synchronously for edge model aggregation, while the ESs forward their edge models to the cloud server semi-asynchronously for global model aggregation. The above training manner leads to a tradeoff between the training loss in each round and the round latency. HPFL combines the objectives of training loss minimization and round latency minimization while jointly determining the optimal bandwidth allocation as well as the ES scheduling policy in the hierarchical learning framework. Extensive experiments verify that HPFL not only guarantees convergence in hierarchical aggregation frameworks but also has advantages in round training loss maximization and round latency minimization.
\end{abstract}


\IEEEpeerreviewmaketitle

\section{Introduction} \label{sec:1}

\IEEEPARstart{F}{ederated} Learning (FL) is a distributed machine learning paradigm that enables multiple mobile user equipments (UEs) to jointly train a common model without ever uploading their raw data to a central parameter server (PS)~\cite{mcmahan2017communication}.
FL has already shown its great potential in exploring behavior patterns hidden in the large amount of data generated by UEs, such as intelligent vehicles, smartphones, and wearable devices.
These patterns, also known as model parameters, will provide guidance to newcomers, thus fostering significant applications such as Artificial Intelligence (AI) medical diagnosis~\cite{rieke2020future} and autonomous vehicles~\cite{xiao2021vehicle}.
Despite the tremendous benefits FL brings to edge intelligence, model training in \emph{heterogenous} and potentially \emph{massive} mobile edge computing (MEC) networks introduces novel challenges that require a fundamental departure from conventional approaches.

Conventionally, FL algorithms are implemented in a two-layer framework with a central PS and multiple distributed UEs~\cite{bonawitz2019towards, zhang2021survey}.
Using the on-device datasets, UEs train local models in reference to the current global model, which are then collected by the PS to further improve the global model.
The PS will then feed the new model back to UEs for a new round of local updating.
However, this typical two-layer framework does not scale well to large-scale FL settings, which are inevitable over massive MEC networks.
This is because, (i) the communication and computation resources in a MEC system are limited~\cite{wang2019adaptive, liu2021resource}. Direct communications between UEs and the remote PS will slow down the overall training due to the scarce resources allocated to each UE;
(ii) the number of UEs ranges from thousands to billions. Such a humongous number of UEs makes it hard for the PS to manage and monitor the training process~\cite{zhang2020enabling, zhan2020incentive}.
Therefore, a two-layer framework is not enough, and a multi-layer FL framework emerges as a  pragmatic solution for large-scale FL implementation in practice~\cite{wang2021resource, chen2021semi, dinh2021enabling}.

In addition, datasets generated by UEs are characterized by a high degree of diversity~\cite{ghosh2019robust, zhao2018federated, diao2020heterofl}. Not only does the number of data samples generated by UEs vary, but these data samples are not independent and identically distributed (non-i.i.d). Learning from such heterogeneous data is difficult, as conventional FL algorithms usually develop a \emph{common} model for all UEs such that the obtained global model could perform arbitrarily poorly on a specific UE. This situation worsens with the rapid development of end-user devices' sensing and computation capabilities. Therefore, it is desired for the FL algorithm to provide personalized services to individual UEs by conducting adaptations to the global model.

Despite the extensive studies on personalized algorithms of FL, remarkably little is known about their implementations in a hierarchical framework. Specifically, \cite{fallah2020personalized} proposed the first work in providing personalized FL services to UEs, named Personalized Federated Learning (PFL). It is an approach that learns an \emph{initial} model that is good enough for all UEs to start with. Using this initial model, each UE can conduct fast adaptation to its local dataset using only a few data points. As a result, the UEs can enjoy fast personalized models by adapting the global model to local datasets. From PFL, a string of follow-ups~\cite{deng2020adaptive, wu2020personalized, t2020personalized, hanzely2020lower, collins2021exploiting, hu2020personalized} modified, developed and extended PFL. However, their focus has been primarily on a two-layer or non-hierarchical FL framework. It is \emph{unclear} how to implement PFL in a hierarchical framework, considering limited communication and computational resources in MEC networks, and whether PFL can converge with such an implementation.

\begin{figure*}
  \centering
  \includegraphics[width=0.9\linewidth]{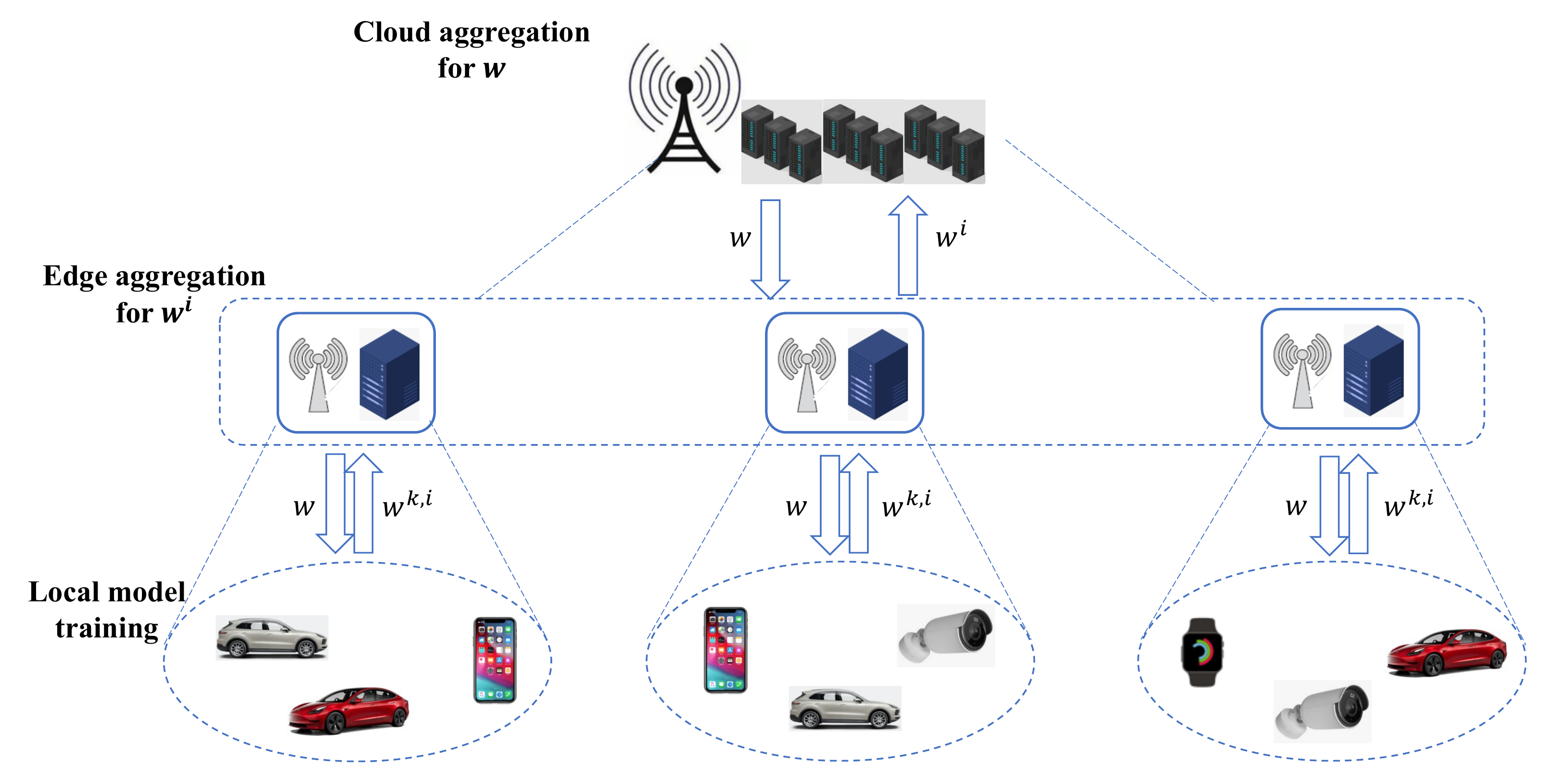}
  \caption{An example of the HPFL framework.}
  \label{fig:hierarchy}
\end{figure*}

In this paper, we propose a PFL algorithm for massive MEC networks, named hierarchical PFL (HPFL), which conducts fast adaptations with hierarchical aggregations.
As illustrated in Fig.~\ref{fig:hierarchy}, HPFL considers a learning framework that consists of three layers, i.e., the UE, edge server (ES), and central server (CS) layers.
Each ES, which serves multiple UEs and can be co-located with small cell base stations, \emph{synchronously} aggregates local models that are trained and sent by UEs.
The CS is connected to a macro base station, \emph{semi-asynchronously} aggregates edge models that are sent by ESs. More specifically, after receiving the current global model, UEs train and update their local models using their datasets. The updated local models are then transmitted to the ES for synchronous edge aggregation. Then, the updated edge models are further transmitted to the CS for semi-asynchronous global model aggregation and update. The updated global model will then be sent back to individual UEs in the same way they came.
A full round of this procedure constitutes one communication round, and the system repeats such interactions until HPFL converges.
With hierarchical aggregations, limited resources are released to serve more UEs, making it much easier for the HPFL scheduler to manage the massive MEC networks.

The semi-asynchronous training across ESs leads to a tradeoff between the per round training loss and the round latency over resource-limited MEC networks.
To that end, our objective is to minimize the weighted sum of the round training loss and the round latency by jointly optimizing ES scheduling policy and bandwidth allocation.
To make the formulated optimization problem tractable, we first analyze the convergence of HPFL. Based on the analysis results, we decompose the optimization problem as a separate ES scheduling problem and bandwidth allocation problem and solve them, respectively. With the obtained results, we can implement HPFL over MEC networks to provide efficient hierarchical training services.

To summarize, in this paper, we make the following contributions:
\begin{itemize}
  \item We develop HPFL, a three-layer FL framework that supports hierarchical aggregations. Specifically, the ES and its UEs conduct synchronous training, while the ESs and CS carry out semi-asynchronous training. By solving a joint ES scheduling and bandwidth allocation problem, HPFL seeks an efficient tradeoff between providing the maximum round training loss and the minimum round latency.

  \item We derive the convergence rate of HPFL for non-convex loss functions. Our analysis shows that the upper bound of the training loss in each round is a linear function of data importance, thus allowing us to reformulate the round training loss maximization problem to the data importance maximization problem.

  \item We solve the reformulated optimization problem by decoupling it into two sub-problems: an ES scheduling problem and a bandwidth allocation problem. Particularly, the ES scheduling problem can be solved by comparing the value of total data importance captured in each round with the value of round latency, while the optimal bandwidth allocation can also be achieved using an online algorithm.

  \item We conduct extensive experiments using the MNIST and CIFAR-10 datasets to verify the effectiveness of HPFL in saving the round latency as well as capturing more important data, compared with intuitive HFL and HPFL algorithms with fully selected or randomly selected ESs.
\end{itemize}

The rest of the paper is organized as follows. In Section~\ref{sec:2} we review two algorithms, hierarchical FL (HFL) and PFL, to introduce some basic ideas and notations that will be used in our proposed algorithm. Then in Section~\ref{sec:3} and Section~\ref{sec:4} we introduce the system model and problem formulation of HPFL. In order to solve the optimization problem of HPFL, we first analyze its convergence in Section~\ref{sec:5}, and then, in Section~\ref{sec:6}, we decompose the problem as a separate ES scheduling problem and a bandwidth allocation problem, and solve them, respectively. At last, we evaluate the performance of HPFL through extensive experiments in Section~\ref{sec:7} and conclude this work in Section~\ref{sec:8}.

\section{Backgrounds} \label{sec:2}

This section reviews the notations and concepts in HFL and PFL algorithms to facilitate the design of our algorithm.

\subsection{Hierarchical Federated Learning} \label{sec:2.1}

Implementing FL in a small area (e.g., companies or hospitals) usually considers a two-tier architecture, where there is a central server and $n$ UEs. The target for the learner is to find the optimal model parameter $\mathbf{w}$ that minimizes the loss function $f(\mathbf{w})=\frac{1}{n}\sum_{i=1}^{n} f_i(\mathbf{w})$ across all UEs, where $f_i(\mathbf{w})$ denotes the loss function of UE $i$ in predicting model parameter $\mathbf{w}$. That is, the goal for all the entities is to obtain $\mathbf{w}^* = \arg\min_{\mathbf{w}} f(\mathbf{w})$, while preserving privacy of all the UEs.

In practice, the optimal $\mathbf{w}$ is attained via $T$ rounds of training across the $n$ UEs. Specifically, at round $t$, after receiving the global model parameter $\mathbf{w}_t$, each UE $i$ uses its local dataset $\mathcal{D}_i$ to update the local model according to  the following
\begin{equation}
  \mathbf{w}_{t+1}^i = \mathbf{w}_t- \beta \nabla f_i(\mathbf{w}_t),
\end{equation}
where $\beta$ denotes the learning rate.
The updated local model parameter $\mathbf{w}_{t+1}^i$ is then uploaded to and aggregated at the central server as $\mathbf{w}_{t+1} = \frac{1}{n} \sum_{i=1}^{n} \mathbf{w}_{t+1}^i$. In this way, the global model progresses from $\mathbf{w}_t$ to $\mathbf{w}_{t+1}$.
This process is repeated until a certain model prediction accuracy $\epsilon$ is achieved.

Regarding large-scale learning tasks, HFL becomes a natural choice for collecting information from a swarm of mobile UEs.
These edge nodes are divided into multiple small areas based on their geographical locations. The small areas are generally separated by tens of kilometers, and each small area contains thousands of UEs.
Therefore, the hierarchical architecture for large-scale FL usually contains at least three tiers (cf. Fig.~\ref{fig:hierarchy}).
Usually, an edge server is in charge of a small area, whereas UEs in this small area will train their local models and then send their updates to the edge server.
The edge server aggregates local models and then updates its edge model.
Formally, if there are $n_k$ UEs in a small area $k$ ($k=1,2,\dots,K$), then the model aggregation at ES $k$ takes the following form
\begin{equation}\label{equ:HFL_edge}
  \mathbf{w}_{t+1}^k = \frac{1}{n_k} \sum_{i=1}^{n_k} \mathbf{w}_{t+1}^{k,i},
\end{equation}
where $\mathbf{w}_{t+1}^{k,i}$ denotes the updated local model at UE $i$, and it is computed by
\begin{equation}
  \mathbf{w}_{t+1}^{k,i} = \mathbf{w}_t^{k,i} - \beta \nabla f_i(\mathbf{w}_t).
\end{equation}

Then, given $K$ small areas for large-scale FL, each ES will send its edge model to the cloud server for global model aggregation, which is computed as follows,
\begin{equation}\label{equ:HFL_cloud}
  \mathbf{w}_{t+1} = \frac{1}{K}\sum_{k=1}^{K}\mathbf{w}_{t+1}^k.
\end{equation}

The main advantage of HFL lies in the significant reduction of communication resources since there exist no direct communications between individual UEs and the cloud server. Updates from individual UEs are uploaded layer by layer, thus greatly reducing the number of entities competing for bandwidth resources simultaneously.

\subsection{Personalized Federated Learning} \label{sec:2.2}

PFL is proposed to address the mismatch problem between local and global models, which is essentially derived from the heterogeneous datasets generated by UEs.
It uses the well-known meta-learning algorithm, MAML, to interpret FL.
The target of PFL is to find a well-performed initial point that is easy for individual UEs to adapt to.
Specifically, PFL aims to find a good initial point, or meta model $\mathbf{w}_{\text{meta}}$. Starting from $\mathbf{w}_{\text{meta}}$, each UE $i$ conducts a fast adaptation to its local dataset with only a few data points using only one or more steps of gradient descent and then obtains its local model, or fine-tuned model, $\mathbf{w}_{\text{fine-tuned}}$.

Overall, the objective of the learner in PFL is to minimize the loss function $F(\mathbf{w})$ across all UEs, where $F_i(\mathbf{w})$ denotes the loss function of UE $i$. Particularly, $F_i(\mathbf{w})$ is defined as follows,
\begin{equation}\label{equ:PFL_loss}
  F_i(\mathbf{w}) = f_i(\mathbf{w} - \alpha \nabla f_i(\mathbf{w})),
\end{equation}
where $\alpha \geq 0$ is the learning rate of individual UEs.
Unlike conventional FL algorithms, after receiving the current global model $\mathbf{w}_t$, a UE in PFL first computes adapted/fine-tuned parameters $\theta_i$ with one step of gradient descent, which is computed as follows,
\begin{equation}\label{equ:theta}
  \theta_i = \mathbf{w}_t-\alpha \nabla f_i(\mathbf{w}_t) .
\end{equation}
Then, based on $\theta_i$, UE $i$ updates its local model parameter as follows:
\begin{equation}
  \mathbf{w}_{t+1}^i = \mathbf{w}_t^i - \beta\nabla_{\mathbf{w}} F_i(\theta_i).
\end{equation}
It is worthwhile to note that the gradient of $F_i(\theta_i)$ contains a Hessian matrix to $\mathbf{w}_t$, i.e.,
\begin{equation}
  \nabla_{\mathbf{w}} F_i(\theta_i) = (I-\alpha \nabla^2 f_i(\mathbf{w}_t)) \nabla f_i(\mathbf{w}_t - \alpha \nabla f_i(\mathbf{w}_t)).
\end{equation}

Finally, at the central server, the global model is updated with the following formula,
\begin{equation}
  \mathbf{w}_{t+1} = \frac{1}{n}\sum_{i=1}^{n}\mathbf{w}_{t+1}^i =\mathbf{w}_t-\beta \frac{1}{n}\sum_{i=1}^{n}\nabla_\mathbf{w} F_i(\theta_i).
\end{equation}

\section{System Model} \label{sec:3}

In this paper, we consider implementing PFL in a hierarchical framework.
With the same hierarchical architecture we introduce in HFL (see Fig.~\ref{fig:hierarchy}),  we consider a set of $K$ ESs $\mathcal{K} = \{1,\dots, K\}$. They can be co-located with small cell base stations~\cite{poularakis2020service}. Each server $k$ has $n_k$ local UEs, denoted as $\mathcal{N}_k = \{1,\dots, n_k\}$. Meanwhile, the edge servers are connected with a macro base station, equipped with a cloud server, denoted by CS $0$. Local models trained by individual UEs are first aggregated at edge servers before finally reaching the cloud server.
We consider \textit{synchronous} FL for the lower level of the hierarchy (i.e., FL between edge servers and mobile UEs) but \textit{semi-asynchronous} FL for the higher level of the hierarchy (i.e., FL between the cloud server and edge servers).
In this section, we introduce the system model following the learning process of model parameter generation and transmission.

\subsection{Local Model Training and Transmission} \label{sec:3.1}

\subsubsection{Local Model Training}

Under each edge server $k\in\mathcal{K}$, local UE $i$ owns a local dataset $\mathcal{D}_{k,i}$, and its size is denoted as $D_{k,i} = |\mathcal{D}_{k,i}|$. Based on $\mathcal{D}_{k, i}$, UE $i$ trains its local model using the iterative formula as follows,
\begin{equation}\label{equ:update_lower}
  \mathbf{w}_{t+1}^{k,i} = \mathbf{w}_{t} - \beta \nabla F_{k,i} (\mathbf{w}_t),
\end{equation}
where $F_{k,i}(\mathbf{w}_t)$ is the PFL loss function defined in (\ref{equ:PFL_loss}) of UE $i$ under edge server $k$ at round $t$. Then, UE $i$ sends $\mathbf{w}_{t+1}^{k,i}$ to its edge server $k$. For simplicity of notation, we consider only one gradient update in the formula and for the rest of the paper, but using multiple local iterations is a straightforward extension. It is also worth noting that we use the full-batch gradient descent to update the local model. In practice, this is not the usual case. Schedulers are encouraged to sample mini-batch of data points on each UE since the volume of data generated by UEs is too large to be processed. While not the focus of this work, the analysis of mini-batch gradient decent can naturally be combined with the methods proposed herein.

As for the computation time of UE $i$ within the coverage of ES $k$, let $c_{k, i}$ denote the number of CPU cycles for UE $i$ to execute the training of one data sample, $\delta_{k,i}$ denote the CPU-cycle frequency of UE $i$.
Then, the computation time of UE $i$ per local iteration can be expressed as follows~\cite{shi2020joint},
\begin{equation}
    \text{Tcmp}_t^{k,i} = \frac{c_{k,i} D_{k,i}}{\delta_{k,i}}.
\end{equation}

\subsubsection{Local Model Transmission}

After UE $i$ updates its local model from $\mathbf{w}_t$ to $\mathbf{w}_{t+1}^{k,i}$, it will send the update to its edge server $k$ for edge model aggregation.
In this paper, we adopt the Orthogonal Frequency Division Multiple Access (OFDMA) technique for local model transmission.
Let $b_t^{k, i}$ be the bandwidth matrix allocated to UE $i$ for its model transmission to edge server $k$ in round $t$, then the achievable uplink rate of UE $i$ to edge server $k$ can be formulated as follows~\footnote{The adoption of SNR (signal-to-noise-ratio) rather than SINR (signal-to-interference-plus-noise-ratio) in this paper omits the consideration of inter-user interference caused by other users that are located in other service areas (ESs other than the considered ES). Nevertheless, it is easy to extend the situation from SNR to SINR. This is because only system parameters that can be measured by channel estimation methods are introduced~\cite{chen2020joint}. Therefore, in this paper we use SNR rather than SINR for simplicity and cleanliness in the following sections.},
\begin{equation} \label{equ:uplink_rate}
  r_t^{k,i} = b_t^{k,i} \log_2\Big(1+\frac{p_{k,i}h_t^{k,i}}{b_t^{k,i}N_0} \Big), \quad \forall k\in\mathcal{K},i\in\mathcal{N}_k,
\end{equation}
where $p_{k,i}$ is the transmit power of UE $i$ to edge server $k$, $h_t^{k,i}$ is the channel gain between UE $i$ and edge server $k$, and $N_0$ is the noise power.

Based on the uplink rate, we are now able to estimate the edge communication time between UE $i$ and server $k$. Let $Z$ be the data size of the model parameter, then the local model upload latency of UE $i$ in round $t$ can be computed by
\begin{equation} \label{equ:tcom}
  \text{Tcom}_t^{k,i} = \frac{Z}{r_t^{k,i}},\quad \forall k\in\mathcal{K},i\in\mathcal{N}_k.
\end{equation}

\subsection{Edge Model Aggregation and Transmission} \label{sec:3.2}

\subsubsection{Edge Model Aggregation}

Once the models from local UEs have arrived, the edge server can aggregate them to update the edge model using the following formula,

\begin{equation}\label{equ:update_upper}
  \mathbf{w}_{t+1}^k = \frac{1}{n_k}\sum_{i=1}^{n_k}\mathbf{w}_{t+1}^{k,i}.
\end{equation}

Note that in this paper, an ES is designed to aggragate local models synchronouly. UEs are randomly selected in each round to transmit their local updates until all selected UEs arrive at the ES for edge model aggragation. This is because the cover range of an ES is small, especially in 5G and B5G networks. For example, the cover range of a BS that supports millimeter Wave (mmWave) communication is usually around 100 m~\cite{mao2021ai}. UEs are sparsely scattered within the coverage of each ES and it is convenient to schedule these UEs synchronously while training.

\subsubsection{Edge Model Transmission}
Since there is no data training on edge servers, the computation time of model aggregation at edge servers is negligible.
As long as $\mathbf{w}_{t+1}^k$ is generated, edge server $k$ will send it to the cloud server for global model aggregation.
Let $b_t^{k,0}$ be the bandwidth allocated to edge server $k$ for its model transmission to the cloud server in round $t$, then the achievable uplink rate of edge server $k$ to the cloud server can be formulated as~\footnote{Here we also use SNR rather than SINR for simplicity and cleanliness in the following sections.}
\begin{equation}
  r_t^{k,0} = b_t^{k,0} \log_2\Big( 1+\frac{p_{k,0}h_t^{k,0}}{b_t^{k,0}N_0} \Big), \quad \forall k\in\mathcal{K},
\end{equation}
where $p_{k,0}$ and $h_{k,0}$ are the transmit power and channel gain of edge server $k$ to the cloud server, respectively.
In consequence, the latency of uploading edge server $k$'s model can be computed by
\begin{equation}
  \text{Tcom}_t^{k,0} = \frac{Z_t^k}{r_t^{k,0}}.
\end{equation}
Note that the data size transmitted by edge server $k$ in round $t$ is denoted as $Z_t^k$, where $Z_t^k\leq Z$. This is because edge servers update the global model semi-asynchronously, and edge server $k$ does not have to transmit the entire model update during round $t$.

\subsection{Cloud Aggregation} \label{sec:3.3}

The cloud server aggregates updates from edge servers semi-asynchronously. That is, in each communication round, once the updates from $A$ ESs arrive at the cloud server, the CS would update the global model. Aggregating edge models semi-asynchronously is the result of the tradeoff between the latency and the training accuracy. On the one hand, if the CS aggregates the updates from ESs synchronously, the latency between ESs and the CS is determined by the latest ES that finishes its data transmission to the cloud. The long distance between ESs and the CS would make this latency unacceptable. On the other hand, if the CS aggregates the updates asynchronously, the learning algorithm may not converge due the staleness caused by those late ESs with out-of-date updates. As a result, for a massive mobile edge network that might suffer unacceptable long round latency as well as learning accuracy, we choose to aggregate edge models semi-asynchronously at the CS.

\begin{figure*}
  \centering
  \includegraphics[width=0.9\linewidth]{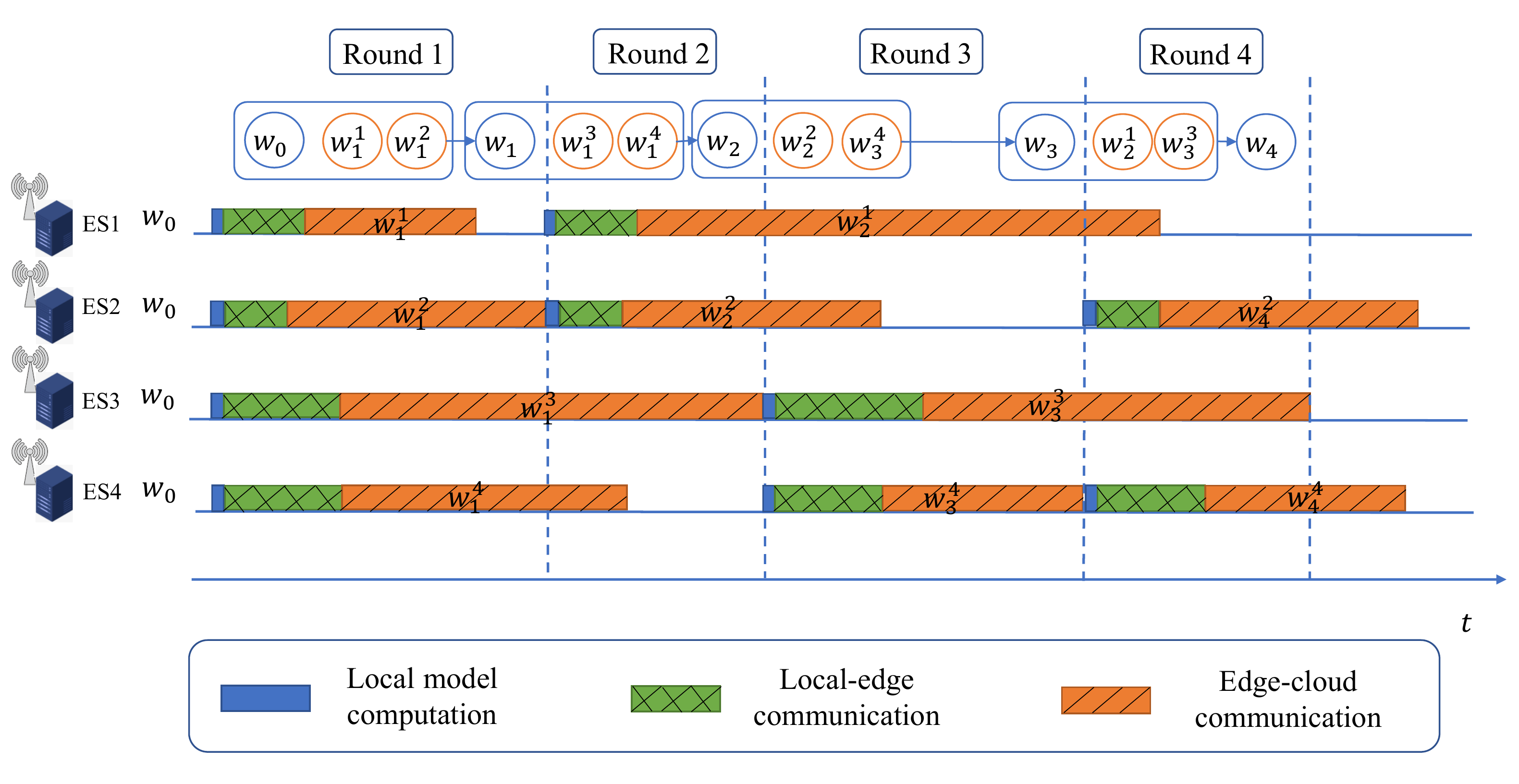}
  \caption{Illustration of the HPFL mechanism, where ESs are scheduled semi-asynchronously. In this example, there are four ESs used for global model aggregation, but only two ESs are selected in each round, that is, $A=2$.}
  \label{fig:scheduling}
\end{figure*}

Given that the cloud server updates the global model in a semi-asynchronous manner, in order to characterize the scheduling policy of the ESs in each round, we introduce an indicator, $\pi_t^k \in \{0,1\}$, to denote whether the edge server $k\in\mathcal{K}$ is selected in round $t$ to update the global model.
Specifically, $\pi_t^k = 1$ means that ES $k$ is selected to update the global model in round $t$, and $\pi_t^k=0$ otherwise.
For instance, consider the example shown in Fig.~\ref{fig:scheduling}, there are four ESs in the system, and only two ESs are selected in each round to updating the global model at the cloud server. The computation has been carried out in 4 rounds and the ES scheduling decision vector  $\mathbf{\Pi}$ in each round can be written as $\{1,1,0,0\}$, $\{0,0,1,1\}$, $\{0,1,0,1\}$ and $\{1,0,1,0\}$, respectively.

Meanwhile, for the edge servers that are not selected in round $t$ but will be selected in later rounds, their updates are stale and may affect the convergence rate. We use $\tau_t^k$ to denote the \emph{interval} between the current round $t$ and the last received global model version by ES $k$. Such an interval reflects the \emph{staleness} of edge updates. With this notion, the edge model received by the cloud server in round $t$ is denoted by $\mathbf{w}_{t-\tau_t^k}^k$. As a result, the global model update at the cloud server in round $t$ can be rewritten as follows,
\begin{equation}\label{equ:global_update}
 \mathbf{w}_{t+1} = \mathbf{w}_t - \frac{\beta}{A} \sum_{k=1}^{K}\frac{\pi_t^k}{n_k} \sum_{i=1}^{n_k} \nabla F_{k,i}(\mathbf{w}_{t-\tau_t^k}),
\end{equation}
where $A = \sum_{k=1}^{K} \pi_t^k$ denotes the number of selected ESs in each round.
Besides, we use $\mathcal{A}_t$ to denote the set of ESs selected to update the global model in round $t$.

At the beginning of the next round $t+1$, the cloud server will send the updated global model $\mathbf{w}_{t+1}$ back to all ESs, which are further forwarded to all UEs.
The downlink broadcasting time consumed by this process is much smaller than the uplink transmission time since the transmit power of upper layers is much higher than that of lower layers, thereby being ignored for simplification~\cite{shi2020joint, you2022semi}.
At this point, we can compute the total round latency of an HPFL system. The total round latency consists of two parts. The first part is the time consumed by the local model update, transmission, and aggregation between UEs and their ES $k$. The duration of this part is determined by the slowest UE within the coverage of ES $k$, which is equal to $\max_{i\in\mathcal{N}_k} \{\text{Tcom}_t^{k,i} + \text{Tcmp}_t^{k,i} \}$.
The second part is the time consumed by the edge model update, transmission, and aggregation between edge servers and the cloud server, which equals $\text{Tcom}_t^{k,0}$.
As a result, the total round latency of ES $k$, which is denoted as $O_t^k$, is computed by
\begin{equation}\label{equ:round_time}
  O_t^k = \max_{i\in \mathcal{N}_k} \{ \text{Tcmp}_t^{k,i}+ \text{Tcom}_t^{k,i} \} + \text{Tcom}_t^{k,0}.
\end{equation}

Note that here we also omit the consideration of aggregation time. This is because compared with the data computation time, the aggregation time is too short to be noticed. At this point, in accordance with the principle of semi-asynchronous learning, the total round latency is determined by the slowest selected ES in that round, i.e., $O_t = \max_{k\in\mathcal{K}} \{\pi_t^k O_t^k\}$.

\section{Problem Formulation} \label{sec:4}

For massive MEC networks, it is desirable to attain a fast convergence for the learning process.
Therefore, in each round, our goal for the cloud server is to aggregate updates from as many ESs as possible. However, the more ESs that are selected, the longer round latency is likely to be.
As a result, we try to find a tradeoff between the amount of information uploaded by the selected ESs in a certain round and the round latency, which is formalized as follows~\footnote{Besides bandwidth allocation and UE scheduling policy, other decision variables such like transmit power can also be included in the problem formulation. The logic keeps the same, but the parameters that need to be considered might change. Problem (P1) shows the case when we consider the bandwidth allocation and UE scheduling policy as variables, and it is free for the researchers to extend this general formulation to other forms.},

\begin{subequations}\label{equ:opt1}
  \begin{align}
    \min_{\mathbf{\Pi}, \mathbf{b}_{k,i}, \mathbf{b}_{k,0}} \quad & -\rho (F(\mathbf{w}_{t+1})-F(\mathbf{w}_{t})) + (1-\rho) \max_{k\in\mathcal{K}} \{\pi_t^k O_t^k\} \label{equ:opt1_obj} \tag{P1} \\
    \text{s.t.} \quad & \pi_t^k = \{0,1\}, \quad \forall k\in\mathcal{K} \label{equ:opt1_const1} \tag{C1} \\
    &  \sum_{k=1}^{K}\sum_{i=1}^{n_k} b_t^{i,k} + \sum_{k=1}^{K} b_t^{k,0} \leq B, \label{equ:opt1_const2} \tag{C2} \\
    & b_t^{k,i}, b_t^{k,0} \geq 0, \quad \forall i\in\mathcal{N}_k, k\in\mathcal{K}, \label{equ:opt1_const3} \tag{C3}
  \end{align}
\end{subequations}
where $\mathbf{b}_{k,i}$ denotes the bandwidth allocation matrix between UE $i\in\mathcal{N}_k$ and ES $k$, while $\mathbf{b}_{k,0}$ denotes the bandwidth allocation vector between ES $k$ and CS $0$. Meanwhile, $\mathbf{\Pi} \triangleq \{\pi_t^1,\pi_t^2,\dots, \pi_t^k\}$ is the ES scheduling decision vector in round $t$.
$\rho \in [0,1]$ is the weight factor that controls the tradeoff between data importance and round latency.
The first constraint (C1) indicates that $\pi_t^k$ is a binary variable. The second constraint (C2) stipulates that the overall bandwidth allocated to the data transmission between UE and ES layers and between ES and CS layers shall not surpass $B$.
The third constraint (C3) shows that the bandwidth allocation is no smaller than $0$.

\section{Convergence Analysis} \label{sec:5}

Before solving the optimization problem, we first analyze the convergence rate of HPFL.
This will offer insights into the key parameters that affect the value of training loss drop (i.e., $F(\mathbf{w}_{t+1}) - F(\mathbf{w}_{t})$)  as well as assisting us in transforming the objective of the HPFL optimization problem (P1) to a tractable form.

\subsection{Assumptions} \label{sec:5.1}

In this paper, we consider the \emph{non-convex} loss functions. To make the convergence analysis consistent with that of PFL, we make the following assumptions~\cite{fallah2020personalized}.

\begin{assum}[Bounded Staleness] \label{assum:1}
  All delay variables $\tau_t^k$'s are bounded, i.e., $\max_{k\in\mathcal{K}} \tau_t^k \leq S$.
\end{assum}
\begin{assum}\label{assum:2}
  For each UE $i\in\mathcal{U}_k$, $k\in\mathcal{K}$, its gradient $\nabla f_{k,i}$ is $L$-Lipschitz continuous and is bound by a nonnegative constant $C$, namely,
  \begin{align}\label{equ:assum2}
    \|\nabla f_{k,i}(\mathbf{w}) - \nabla f_{k,i} (\mathbf{u})\| & \leq  L\|\mathbf{w} - \mathbf{u}\|, \quad \mathbf{w},\mathbf{u}\in\mathbb{R}^m, \\
    \|\nabla f_{k,i}(\mathbf{w})\| & \leq C, \quad \mathbf{w}\in\mathbb{R}^m.
  \end{align}
\end{assum}
\begin{assum}\label{assum:3}
  For each UE $i\in\mathcal{U}_k$, $k\in\mathcal{K}$, its Hessian is $\rho$-Lipschitz continuous, that is,
  \begin{equation}\label{equ:assum3}
    \|\nabla^2 f_{k,i}(\mathbf{w}) - \nabla^2 f_{k,i}(\mathbf{u}) \| \leq \rho \|\mathbf{w}-\mathbf{u} \|, \quad \mathbf{w},\mathbf{u}\in\mathbb{R}^m.
  \end{equation}
\end{assum}
\begin{assum}\label{assum:4}
  For any $\mathbf{w}\in\mathbb{R}^m$, the gradient and Hessian of local loss function $f_{k,i}(\mathbf{w})$ and the average loss function $f(\mathbf{w}) = \frac{1}{K}\sum_{k\in\mathcal{K}}\frac{1}{n_k} \sum_{i=1}^{n_k} f_{k,i}$ satisfy the following conditions,
  \begin{align}\label{equ:assum4}
    \frac{1}{K}\sum_{k=1}^K\frac{1}{n_k}\sum_{i=1}^{n_k} \|\nabla f_{k,i}(\mathbf{w}) - \nabla f(\mathbf{w}) \|^2 & \leq \gamma_G^2, \\
    \frac{1}{K}\sum_{k=1}^K\frac{1}{n_k}\sum_{i=1}^{n_k} \|\nabla^2 f_{k,i}(\mathbf{w}) - \nabla^2 f(\mathbf{w}) \|^2 & \leq \gamma_H^2,
  \end{align}
  where $\gamma_G, \gamma_H>0$ are two constants that are small enough.
\end{assum}

While Assumption 1 limits the maximum of the ES staleness, Assumptions~\ref{assum:2} to~\ref{assum:4} characterize the properties of the gradient and Hessian of $f_{k,i}(\mathbf{w})$, which are necessary to deduce the following lemmas and convergence rate analysis.

\subsection{Analysis of Convergence Bound} \label{sec:5.2}

We analyze the convergence bound of the HPFL mechanism in this subsection. Our focus is primarily on how the difference between $F(\mathbf{w}_{t+1})$ and $F(\mathbf{w}_t)$ changes in each round by the global model updating in~(\ref{equ:global_update}). Before delving into the details of convergence analysis, we introduce two lemmas inherited from~\cite{fallah2020personalized} to quantify the smoothness of $F_k(\mathbf{w})$ and $F(\mathbf{w})$, and the deviation between $\nabla F_k(\mathbf{w})$ and $\nabla F(\mathbf{w})$, respectively.

\begin{lem} \label{lem:1}
  If Assumptions~\ref{assum:2}-\ref{assum:4} hold, then $F_k(\mathbf{w})$ is smooth with respect to parameter $L_F:= 4L +\alpha \rho C$. As a result, the average function $F(\mathbf{w}) = \frac{1}{K}\sum_{k=1}^{K} F_k(\mathbf{w})$ is also smooth with parameter $L_F$.
\end{lem}

\begin{lem} \label{lem:2}
  If Assumptions~\ref{assum:2}-\ref{assum:4} hold and $\alpha\in(0,1/L]$, then for any $\mathbf{w}\in\mathbb{R}^{m}$, we have
  \begin{equation}
    \frac{1}{K}\sum_{k=1}^{K} \|\nabla F_k(\mathbf{w}) - \nabla F(\mathbf{w})\|^2 \leq \gamma_F^2,
  \end{equation}
  where $\gamma_F^2$ is defined as
  \begin{equation}
    \gamma_F^2:=3C^2\alpha^2\gamma_H^2+192\gamma_G^2,
  \end{equation}
  where $\nabla F(\mathbf{w}) = \frac{1}{K} \sum_{k=1}^{K} \nabla F_k(\mathbf{w})$.
\end{lem}
From Lemma 2, it is easy to verify that
\begin{equation}
  \frac{1}{n_k} \sum_{i=1}^{n_k} \|\nabla F_{k,i} (\mathbf{w}) - \nabla F_k(\mathbf{w})\|^2 \leq \gamma_F^2.
\end{equation}

Based on the above two lemmas, we can obtain the following theorem.

\begin{thm} \label{thm:1}
  If Assumptions~\ref{assum:1}-\ref{assum:4} hold and the global step size $\beta = 1/L_F$, the upper bound of the difference between $F(\mathbf{w}_{t+1})$ and $F(\mathbf{w}_t)$ is
  \begin{equation}\label{equ:thm1.1}
    F(\mathbf{w}_{t+1}) - F(\mathbf{w}_t) \leq \phi \sum_{k=1}^{K} \pi_t^k \| \nabla F (\mathbf{w}_{t-\tau_t^k})\|^2 + \nu,
  \end{equation}
  where $\nu$ and $\phi$ are two constants such that
  \begin{align}\label{equ:thm1.2}
    \phi = & \frac{5 \beta S^2}{A}, \nonumber \\
    \nu = & \frac{10 \beta K\gamma_F^2}{A} +
  \frac{5\beta S^2K \gamma_F^2}{A}.
  \end{align}
\end{thm}

\begin{IEEEproof}
See the appendix.
\end{IEEEproof}


From Theorem~\ref{thm:1}, we obtain the upper bound of $F(\mathbf{w}_{t+1}) - F(\mathbf{w}_t)$, and the objective to minimize $F(\mathbf{w}_{t+1}) - F(\mathbf{w}_t)$ can be approximated to minimize its upper bound. If we ignore the constant terms in the upper bound, the original objective function of the optimization problem (P1) can be recast as

\begin{equation} \label{equ:obj_transformed}
  \min_{\mathbf{\Pi},\mathbf{b}_{k,i},\mathbf{b}_{k,0}} \quad -\rho \phi \sum_{k\in\mathcal{K}} \pi_t^k \|\nabla F(\mathbf{w}_{t-\tau_t^k})\|^2
 + (1-\rho) \max_{k\in\mathcal{K}}\{\pi_t^k O_t^k\}
\end{equation}

Using this transformed objective, we can optimize values of $\mathbf{\Pi}$, $\mathbf{b}_{k,i}$, and $\mathbf{b}_{k,0}$ in the HPFL optimization problem (P1).

Notably, the term that matters in the upper bound, $\phi\sum_{k=1}^{K} \pi_t^k \|\nabla F(\mathbf{w}_{t-\tau_t^k})\|^2$, is the secondary moment of local gradient-norm-value (GNV) of ES $k$. This form of GNV is widely adopted as a measure of \emph{data importance} indicating the amount of information provided by the local datasets~\cite{chen2021semi}. Therefore, the transformed objective function (\ref{equ:obj_transformed}) can be regarded as the tradeoff between the total data importance aggregated by the CS in each round and the round latency.

\section{Joint Edge Server Scheduling and Bandwidth Allocation} \label{sec:6}

This section details the steps to solve the optimization problem (P1). In particular, we decouple (P1) into two subproblems, the ES selection problem and bandwidth allocation problem, and then solve the two subproblems separately.

\subsection{Problem Decoupling} \label{sec:6.1}

Using the results from the convergence analysis, the optimization problem (P1) of HPFL can be approximatively transformed into the following optimization problem,
\begin{subequations}\label{equ:opt2}
  \begin{align}
  \min_{\mathbf{\Pi},\mathbf{b}_{k,i},\mathbf{b}_{k,0}} \text{  } & -\rho \phi \sum_{k\in\mathcal{K}} \pi_t^k \|\nabla F(\mathbf{w}_{t-\tau_t^k})\|^2 \nonumber \\
  & + (1-\rho) \max_{k\in\mathcal{K}}\{\pi_t^k O_t^k\} \label{equ:opt2_obj} \tag{P2} \\
  \text{s.t.} \quad & (\text{C}1) - (\text{C}3). \nonumber
  \end{align}
\end{subequations}
In order to solve (P2), we use the Coordinate Descent~\cite{wright2015coordinate} method.
To be more concrete, in each round, we first consider the bandwidth allocation $\mathbf{b}_{k,i}$, $\mathbf{b}_{k,0}$ as constants and then compute the optimal edge server scheduling policy.
Then, we use the obtained edge server selection $\mathbf{\Pi}$ to compute the bandwidth allocation that can be used in the next round. As a result, we decouple the optimization problem (P2) into two subproblems: edge server selection problem (P3) and bandwidth allocation problem (P4). For a given set of bandwidth, the edge server selection problem can be formalized as follows,
\begin{subequations}\label{equ:opt3}
  \begin{align}
   \min_{\mathbf{\Pi}} \quad & -\rho\phi \sum_{k\in\mathcal{K}} \pi_t^k \|\nabla F(\mathbf{w}_{t-\tau_t^k})\|^2 + (1-\rho) \max_{k\in\mathcal{K}} \{\pi_t^k O_t^k\}. \label{equ:opt3_obj} \tag{P3} \\
   \text{s.t.} \quad & (\text{C}1).\nonumber
  \end{align}
\end{subequations}
With a given ES selection policy, the bandwidth allocation problem is formalized as follows,
\begin{subequations}\label{equ:opt4}
  \begin{align}
    \min_{\mathbf{b}_{k,i}, \mathbf{b}_{k,0}} \quad & \max_{k\in\mathcal{K}} \{\pi_t^k O_t^k\} \label{equ:opt4_obj} \tag{P4} \\
    \text{s.t.} \quad & (\text{C}2)-(\text{C}3).\nonumber
  \end{align}
\end{subequations}

\subsection{Edge Server Scheduling} \label{sec:6.2}

To solve (P3), we introduce an auxiliary variable $Q$ such that $Q = \max_{k\in\mathcal{K}} -\rho \sum_{k\in\mathcal{K}} \pi_t^k \|\nabla F(\mathbf{w}_{t-\tau_t^k})\|^2 + (1-\rho)\pi_t^k O_t^k$. Then, the optimization problem (P3) becomes,
\begin{subequations}\label{equ:opt5}
  \begin{align}
    \min_{\mathbf{\Pi}, Q} \quad & Q \label{equ:opt5_obj} \tag{P5} \\
    \text{s.t.} \quad & -\rho \phi \sum_{k\in\mathcal{K}} \pi_t^k \|\nabla F(\mathbf{w}_{t-\tau_t^k})\|^2 + (1-\rho)\pi_t^k O_t^k \leq Q \label{equ:opt5_const1} \tag{C4} \\
    & (\text{C}1). \nonumber
  \end{align}
\end{subequations}

To solve this problem, we first transform (C1) to its equivalent form, which is the intersection of the following regions,
\begin{align}
  & 0 \leq \pi_t^k \leq 1,\quad \forall k\in\mathcal{K}, \label{equ:opt2_const3} \tag{C5}\\
  & \pi_t^k - (\pi_t^k)^2 \leq 0, \quad \forall k\in\mathcal{K}. \label{equ:opt2_const4} \tag{C6}
\end{align}

At this point, (P5) is a continuous optimization problem with constraints (C4) $-$ (C6) with respect to $\mathbf{\Pi}$. Our goal, however, is to obtain integer solutions for $\pi_t^k$. To achieve this goal, we add a cost function to the objective of (P5) to penalize it if the values of $\pi_t^k$'s are not integers. Thus, (P5) is modified to
\begin{align}
  \min_{\mathbf{\Pi}, Q} \quad & L(\mathbf{\Pi},\lambda)
  \label{equ:opt6_obj} \tag{P6} \\
  \text{s.t.} \quad & (\text{C4}), (\text{C5}), \nonumber
\end{align}
where $L(\mathbf{\Pi},\lambda)$ is the Lagrangian of (P5), given by
\begin{equation}
  L(\mathbf{\Pi},\lambda) \triangleq Q - \sum_{k\in\mathcal{K}} \lambda_k (\pi_t^k - (\pi_t^k)^2),
\end{equation}
where $\lambda_k$ is the penalty factor, and $\lambda_k \geq 0$. It is easy to find that (P6) is a convex optimization problem. This is because the objective is the summation of an affine function of $Q$ and a convex function of $\pi_t^k$, and the feasible domain defined by the constraints of (P4) is linear.

In order to solve (P4), we first transform the constrained optimization problem (P4) into an unconstrained optimization problem. That is, we complete the Lagrangian of (P4), which is obtained as follows,
\begin{align}
  & L(\pi_t^k, Q, \lambda_k, \vartheta) \nonumber \\
  = & Q - \sum_{k\in\mathcal{K}}\lambda_k (\pi_t^k - (\pi_t^k)^2) \nonumber \\
  + & \vartheta ( \rho \phi \sum_{k\in\mathcal{K}} \pi_t^k \|\nabla F(\mathbf{w}_{t-\tau_t^k})\|^2 + (\rho-1)\pi_t^k O_t^k + Q)
\end{align}
where $\lambda$ and $\vartheta$ are the Lagrangian multipliers.
At this point, we can obtain the KKT equation set of (P4), that is,
\begin{align}
\left\{
\begin{aligned}
  & \frac{\partial L(\pi_t^k, Q, \lambda_k, \vartheta)}{\partial \pi_t^k}, \frac{\partial L(\pi_t^k, Q,\lambda_k, \vartheta)}{\partial Q} = 0 \\
  & (\text{C}2), (\text{C}4)-(\text{C}6) \\
  & \lambda_k \geq 0, \quad \forall k\in\mathcal{K} \\
  & \vartheta \leq 0  \\
  &\sum_{k\in\mathcal{K}} \lambda_k (\pi_t^k - (\pi_t^k)^2) =0 \\
  & \vartheta (-\rho \phi \sum_{k\in\mathcal{K}} \pi_t^k \|\nabla F(\mathbf{w}_{t-\tau_t^k})\|^2 + (1-\rho)\pi_t^k O_t^k - Q) = 0
\end{aligned}
\right.
\end{align}
Specifically, the first-order partial derivatives of $L(\pi_t^k, Q, \lambda_k, \vartheta)$ with respect to $\pi_t^k$ and $Q$ are computed by the following equations, respectively,
\begin{align}
  \left\{
  \begin{aligned}
    \frac{\partial L(\pi_t^k, Q, \lambda_k, \vartheta)}{\partial \pi_t^k} = & -\lambda_k(1-2\pi_t^k) + \vartheta\rho\phi\|\nabla F(\mathbf{w}_{t-\tau_t^k})\|^2 \\
    & + \vartheta(\rho-1)O_t^k \\
    \frac{\partial L(\pi_t^k, Q,\lambda_k, \vartheta)}{\partial Q} = & 1 + \vartheta
  \end{aligned}
  \right.
\end{align}
Solving the above equations, we obtain the following results,
\begin{subequations}
\begin{align}
  \vartheta & = -1, \label{equ:39} \\
  \pi_t^k & = \left \{
\begin{aligned}\label{equ:40}
  & \frac{1}{2} + \frac{\rho\phi\|\nabla F(\mathbf{w}_{t-\tau_t^k})\|^2 - (1-\rho)O_t^k}{2\lambda_k} , \quad \text{if } \lambda_k > 0, \\
  & \{0,1\},\quad \text{if } \lambda_k = 0, \rho\phi\|\nabla F(\mathbf{w}_{t-\tau_t^k})\|^2 = (1-\rho)O_t^k.
  \end{aligned}
  \right.
\end{align}
\end{subequations}

Our next step is to solve the value of $\pi_t^k$ according to (\ref{equ:40}) and the integer constraints of $\pi_t^k$ (C5) and (C6). When $\lambda_k> 0$, let $P_k = \rho\phi\|\nabla F(\mathbf{w}_{t-\tau_t^k}) \|^2 - (1-\rho)O_t^k$, then we have $\pi_t^k = \frac{1}{2} + \frac{P_k}{2\lambda_k}$. $P_k$ can be regarded at a constant when the bandwidth allocation is settled. Applying (\ref{equ:40}) to (C5) and (C6), we have
\begin{equation}\label{equ:41}
  \lambda_k = \{-P_k,P_k\}.
\end{equation}
Given that $\lambda_k > 0$, we need to discuss the value of $P_k$ in the following two case:
\begin{itemize}
  \item $\rho\phi\|\nabla F(\mathbf{w}_{t-\tau_t^k}) \|^2 > (1-\rho)O_t^k$. In this case, $P_k>0$. Then, according to (\ref{equ:41}), we have $\lambda_k = P_k$. As a result, based on the equation (\ref{equ:40}), we have $\pi_t^k = 1$;
  \item $\rho\phi\|\nabla F(\mathbf{w}_{t-\tau_t^k}) \|^2 < (1-\rho)O_t^k$. In this case, $P_k<0$. Then, according to (\ref{equ:41}), we have $\lambda_k = -P_k$. As a result,  based on the equation (\ref{equ:40}), we have $\pi_t^k = 0$.
\end{itemize}

Then, when $\lambda_k = 0$, according to the second equation shown in (\ref{equ:40}), we have $\rho\phi\|\nabla F(\mathbf{w}_{t-\tau_t^k}) \|^2 =(1-\rho)O_t^k$, and $\pi_t^k$ can be randomly equal to $0$ or $1$.

From the above analysis, we find that it is easy to determine the optimal value of $\pi_t^k$, as long as we obtain the values of $\rho\|\nabla F(\mathbf{w}_{t-\tau_t^k}) \|^2$ and $(1-\rho)O_t^k$. For a given set of bandwidth allocation, what we need to do is to compute and compare the above two values for each of the ESs, and then determine whether an ES is to be scheduled in this round or not. This makes it easy and fast for practical implementation of ES selection. The pseudo-code of the ES scheduling algorithm is detailed in Alg.~\ref{alg:ES_selection}, and it is named \texttt{ESScheduling}. The \texttt{ESScheduling} algorithm is deployed at the CS since it requires global information (i.e., the global model and the round latency). Meanwhile, it requires $O(K)$ work every time it is triggered.

\begin{algorithm}[t]
\caption{\texttt{ESScheduling}$(\mathbf{b}_{k,i}, \mathbf{b}_{k,0})$}
\label{alg:ES_selection}
\DontPrintSemicolon
\For{$k=1$ \KwTo $K$}{
    $O_{t}^k:= \max_{i\in\mathcal{N}_k}\{\text{Tcmp}_{t}^{k,i} + \text{Tcom}_{t}^{k,i}\} + \text{Tcom}_{t}^{k,0}$ \;
   \eIf{$\rho\phi\|\nabla F(\mathbf{w}_{t-\tau_t^k}) \|^2 \geq (1-\rho)O_t^k$}{
    $\pi_t^k = 1$
   }{$\pi_t^k = 0$}
}
\end{algorithm}

\subsection{Bandwidth Allocation} \label{sec:6.3}

After obtaining the optimal edge server scheduling policy $\mathbf{\Pi}^*$, we proceed to solve the bandwidth allocation problem (P4).
We introduce another auxiliary variable $G$, such that $G = \max_{i\in\mathcal{N}_k} \{ \text{Tcom}_t^{k,i} + \text{Tcmp}_t^{k,i}\}$. Therefore, the bandwidth allocation problem (P4) becomes,
\begin{subequations}\label{equ:opt7}
\begin{align}
  \min_{\mathbf{b}_{k,i},\mathbf{b}_{k,0}, G} \quad & \max_{k\in\mathcal{K}} \{\pi_t^k (G + \text{Tcom}_t^{k,0}) \}
  \label{equ:opt7_obj} \tag{P7} \\
  \text{s.t.} \quad & \text{Tcom}_t^{k,i} + \text{Tcmp}_t^{k,i}\leq G, \quad \forall k\in\mathcal{K} \label{equ:opt7_const1} \tag{C7} \\
  & (\text{C}2) - (\text{C}3). \nonumber
\end{align}
\end{subequations}
$G$ can be regarded as the round latency between a edge server $k$ and its local UEs. We introduce the following theorem to explore relationship between $b_t^{k,i}$ and $G$.
\begin{thm}\label{thm:2}
    When an edge server updates its edge model $\mathbf{w}_t^k$ after receiving $n_k$ gradients from all its UEs, the optimal bandwidth allocation at the edge $\mathbf{b}_t^{k,i}$ can be achieved if and only if all these UEs have the same finishing time. That is, the optimal edge bandwidth allocation $b_t^{k,i}$ between edge server $k\in\mathcal{K}$ and the UEs under its coverage is achieved if and only if $G_1 = G_2 = \dots = G_{N_k} = G$, where $G_i= \text{Tcom}_t^{k,i} + \text{Tcmp}_t^{k,i}$ is the round latency between edge server $k$ and one of its UE $i$, and $G = \max_{i\in\mathcal{N}_k} G_i$.
\end{thm}
\begin{IEEEproof}
Recall the expression of the edge uplink rate $r_t^{k,i}$ defined in (\ref{equ:uplink_rate}), its first-order derivative with respect to $b_{t}^{k,i}$ is computed as follows,
\begin{align}
  \frac{\partial r_t^{k,i}}{\partial b_t^{k,i}} = & \frac{\partial}{\partial b_t^{k,i}} \left(b_t^{k,i} \log_2 \left(1+\frac{p_{k,i}h_t^{k,i}}{b_t^{k,i} N_0}\right)\right)\nonumber \\
  = & \log_2 \left( 1 + \frac{p_{k,i} h_t^{k,i}}{b_t^{k,i} N_0}\right) - \frac{p_{k,i}h_t^{k,i}}{b_t^{k,i}N_0 + p_{k,i}h_t^{k,i}} \nonumber \\
  > & \frac{\frac{p_{k,i} h_t^{k,i}}{b_t^{k,i} N_0}}{1+\frac{p_{k,i} h_t^{k,i}}{b_t^{k,i} N_0}} - \frac{p_{k,i}h_t^{k,i}}{b_t^{k,i}N_0 + p_{k,i}h_t^{k,i}} \nonumber \\
  = & 0,
\end{align}
where the inequality is derived from the fact that $\log_2 (1+x) > \frac{x}{1+x}$. Therefore, $r_t^{k,i}$ monotonically increases with $b_t^{k,i}$.
Given the relationship between $r_t^{k,i}$ and $\text{Tcom}_t^{k,i}$ shown in (\ref{equ:tcom}), it is clear that the Tcom$_t^{k,i}$ is monotonically decreases with $b_t^{k,i}$.
As a result, at round $t$, if any UE $i \in\mathcal{N}_k$ has finished its local model update process faster than the other UEs under the coverage of the same edge server $k$, we can decrease its bandwidth allocation to make up for the other slower UEs $j$ ($j\in\mathcal{N}_k, j\neq i$). The decrease of $b_t^{k,i}$ leads to the increase of Tcom$_t^{k,i}$, and further, to the increase of $G_i$. Meanwhile, the increase of $b_t^{k,j}$ leads to the decrease of Tcom$_t^{k,j}$, and further, to the decease of $G_j$. In this way, $G$ can be reduced since it is determined by the slowest UE in $\mathcal{N}_k$. Such a bandwidth compensation is performed until all UEs in $\mathcal{N}_k$ finish their data transportation with edge server $k$ at the same time. That is, the optimal bandwidth allocation $b_t^{k,i}$ between UE $i\in\mathcal{N}_k$ and edge server $k$ is achieved when all UEs have the same $G_i$, and $G = G_i$.
\end{IEEEproof}

Let $b_t^k = \sum_{i=1}^{n_k} b_t^{k,i}$, then according to~\cite{shi2020joint}, the optimal bandwidth allocation of $b_t^{k,i}$ is computed as follows,
\begin{equation} \label{equ:lower_bandwidth}
  b_t^{k,i} = b_t^k \frac{Z\ln 2}{(G^*(\Pi)-\text{Tcmp}_t^{k,i})(W(-\Gamma_t^{k,i} e^{-\Gamma_t^{k,i}}) + \Gamma_t^{k,i})},
\end{equation}
where $\Gamma_t^{k,i} \triangleq \frac{N_0 Z \ln 2}{(G^*(\Pi) - \text{Tcmp}_t^{k,i})p_{k,i}h_t^{k,i}}$, $W(\cdot)$ is Lamber-W function, and $G^*(\Pi)$ is the objective value of (P7).

Meanwhile, from Theorem~\ref{thm:2}, we can immediately draw the following corollary.
\begin{cor}
  $\text{Tcom}_t^{k,0}$ monotonically decreases with $b_t^{k,0}$. That is, as for the communication between edge servers and the cloud server, the optimal bandwidth allocation can only be achieved if and only if all the scheduled edge servers have the same finishing time.
\end{cor}

At this moment, the relationship between $b_t^k$, $b_t^{k,i}$ and the optimal value $G^*(\mathbf{\Pi})$ is clear. Similarly, the relationship between $b_t^{k,0}$ and $\text{Tcom}_t^{k,0}$ is clear. As a result, (P7) can be regarded as an optimization problem with respect to $G^*$ and $\text{Tcom}_t^{k,0}$. This is a typical min-max optimization problem, which can be solved using the well-known progressive filling~\cite{bertsekas2021data} algorithm.
More specifically, in order to solve the optimization problem (P7), we increase the value of $G$ and $\text{Tcom}_t^{k,0}$ at the same speed until (C2), the bandwidth constraint, has been violated.
To translate this process with respect to $b_t^{k,i}$ and $b_t^{k,0}$, what we need to do to solve (P7) is to increase $b_t^{k,i}$ and $b_t^{k,0}$ at a rate such that $G$ and $\text{Tcom}_t^{k,0}$ can increase at the same speed until the available bandwidth $B$ has been fully occupied. The pseudo-code of the bandwidth allocation mechanism is detailed in Alg.~\ref{alg:bandwidth_alloc}, and it is named \texttt{BandwidthAllocation}. The \texttt{BandwidthAllocation} algorithm is also deployed at the CS since it requires global information to have the bandwidth between ESs and its UEs as well as the bandwidth between the CS and the ESs allocated. Given that \texttt{BandwidthAllocation} uses the progressive filling method, it requires to sort the current bandwidth of the UEs and the ESs. Therefore, the computation complexity of \texttt{BandwidthAllocation} can be coarsely regarded as $O( \log K)$~\cite{you2022hierarchical}.

\begin{algorithm}[t]
\caption{\texttt{BandwidthAllocation}$(\mathbf{\Pi})$}
\label{alg:bandwidth_alloc}
\DontPrintSemicolon
\SetKwFunction{greedy}{ProgressiveFilling}
\SetKwProg{Pro}{Procedure}{}{}
$\mathbf{b}_k, \mathbf{b}_{k,0} = \texttt{ProgressiveFilling}(\Pi)$ \;
Compute $b_t^{k,i}$ with~(\ref{equ:lower_bandwidth})\;
\Pro{\greedy{}}{
    $\begin{aligned}
        \max \qquad & O, \\
        s.t. \qquad & \pi_t^k O_t^k \geq O, \quad \forall k\in\mathcal{K}, \\
                    & \sum_{k\in\mathcal{K}} b_t^k + b_t^{k,0} \leq B,\\
    \end{aligned}$\;
    \KwRet $(\mathbf{b}_{k}, \mathbf{b}_{k,0})$
    }
\end{algorithm}

\subsection{The HPFL algorithm} \label{sec:6.4}

With the obtained ES scheduling policy and bandwidth allocation mechanism, we can now implement HPFL over MEC networks. Alg.~\ref{alg:HPFL} shows the pseudo-code of HPFL. In each communication round, HPFL first computes the bandwidth allocation $\mathbf{b}^{k,0}$ and $\mathbf{b}^{k,i}$ with the ES scheduling vector $\mathbf{\Pi}$ derived from the last round. Then, based on the bandwidth allocation, each ES can decide whether it needs to be scheduled in the current round or not. With the obtained scheduling policy $\mathbf{\Pi}$, HPFL determines the set of ESs contributing to the global model update in this round. Other edge models must be patient until they are called in the next rounds, whether their updates are fresh or stale.

\begin{algorithm}[t]
\caption{HPFL}
\label{alg:HPFL}
\DontPrintSemicolon
Initial: $\mathbf{w}_0$, $\mathcal{A}_t = \varnothing$, $A$, $\pi_{-1}^k =1 (\forall k\in\mathcal{K})$ \;
\For{$t=0$ \KwTo $T-1$}{
    $b_t^{k,0}, b_t^{k,i} = \texttt{BandwidthAllocation}(\{\pi_{t-1}^k\})$ \;
    \For{$k=1$ \KwTo $K$}{
        $\pi_t^k := \texttt{ESScheduling} ({b}_t^{k,0}, b_t^{k,i})$\;
        \If{$\pi_t^k = 1$ and $A \geq 0$}{
            $\mathcal{A}_t \leftarrow \mathcal{A}_t \cup \{k\}$\;
            $A := A - 1$\;}
        \For{$i=1$ \KwTo $n_k$}{
            $\rhd$ \textbf{Local model training at UEs} \;
            Update local model from $\mathbf{w}_{t}^{k,i}$ to $\mathbf{w}_{t+1}^{k,i}$ according to Eq.~(\ref{equ:update_lower})\;
        }
    $\rhd$ \textbf{Edge model aggregation at ESs} \;
    Receive local model $\mathbf{w}_{t+1}^{k,i}$ from UE $i\in \mathcal{N}_k$ \;
    Update edge model from $\mathbf{w}_{t}^k$ to $\mathbf{w}_{t+1}^k$ according to Eq.~(\ref{equ:update_upper}) \;
    }
$\rhd$ \textbf{Global model aggregation at the CS} \;
Receive edge model $\mathbf{w}_{t+1}^k$ from selected ES set $\mathcal{A}_t$\;
Update global model from $\mathbf{w}_t$ to $\mathbf{w}_{t+1}$ according to Eq.~(\ref{equ:global_update}) \;
Send $\mathbf{w}_{t+1}$ back to all UEs\;
}
\end{algorithm}

\section{Performance Evaluation} \label{sec:7}

In this section, we conduct experiments to (i) verify that HPFL can provide convergent training loss, (ii) demonstrate that there exists a tradeoff between the maximum round data importance and minimum round latency in HPFL, and (iii) check the price HPFL pays for its gain in saving the round latency as well as its benefits in capturing more important data.

\subsection{Setup} \label{sec:7.1}

\subsubsection{Datasets and Models}

\begin{table}
\centering
\caption{System Parameters}
\label{tab:sys_para}
\begin{tabular}{|c|c|}
  \hline
  \textbf{Parameter} & \textbf{Value} \\
  \hline
  $\alpha$ (MNIST) & $0.03$ \\
  \hline
  $\beta$ (MNIST) & $0.07$ \\
  \hline
  $\alpha$ (CIFAR-10)& $0.02$\\
  \hline
  $\beta$ (CIFAR-10)& $0.06$ \\
  \hline
  $B$ & 5 MHz \\
  \hline
  $N_0$ & $-174$ dBm/Hz \\
  \hline
  $p_{k,i}$ & $0.01$ W \\
  \hline
  $c_{k,i}$ & $20$ cycles/bit \\
  \hline
  $\delta_{k,i}$ & $2$ GHz \\
  \hline
\end{tabular}
\end{table}

We consider a hierarchical system with one central cloud server and 20 edge servers, each covering an area with 10 UEs.
It is desired for the learner to pick up $15$ ESs in each round, that is, $A=15$.
We consider the channel gains $h_t^{k,i}=o_id_{k,i}^{-2}$ and $h_t^{k,0}=o_kd_{k,0}^{-2}$ follow the Rayleigh distribution where $o_i=-36$dB and $o_k=-40$dB are the Rayleigh fading parameters, $d_{k,i}$ is the actual distance between UE $i$ and ES $k$ that is uniformly generated between $[2, 50]$ m, and $d_{k,0}$ is the actual distance between ES $k$ and BS $0$ that is uniformly generated between $[50, 200]$ m.
We conduct the experiments using two datasets: MNIST~\cite{MNIST} and CIFAR-10~\cite{CIFAR10}. The network model for MNIST is a 2-layer deep neural network (DNN) with a hidden layer of size 100, while the network model for CIFAR-10 is LeNet-5~\cite{lecun1998gradient} that has two convolutional layers and three fully connected layers. The other parameters we use are summarized in Table~\ref{tab:sys_para}. All experiments are conducted by PyTorch~\cite{paszke2019pytorch} version 1.11.0.

\subsubsection{Dataset Participation}

The level of divergence in the distribution of UEs' datasets will affect the overall performance of the system. To reflect this feature, each UE is allocated a different local data size and has $l=1,2\dots, 10$ of the 10 labels. where $l$ indicates the level of data heterogeneity. The higher $l$ is, the more heterogenous the datasets are.

\subsubsection{Baselines}

To examine the effectiveness of HPFL, we compare it with one baseline algorithm in different ES selection modes. Specifically, we compare HPFL with the Hierarchical Federated Learning (HFL) algorithm when ESs are selected in the mode we proposed in this paper, fully selected and randomly selected.
In brief, we compare the performance of HPFL with five other algorithms, namely, HFL with the proposed ES selection mode, HFL with fully selected ESs, HFL with randomly selected ESs, HPFL with fully selected ESs, and HPFL with randomly selected ESs.

\begin{figure*}[!t]
  \centering
  \subfloat[MNIST training loss]{
      \includegraphics[width=3in]{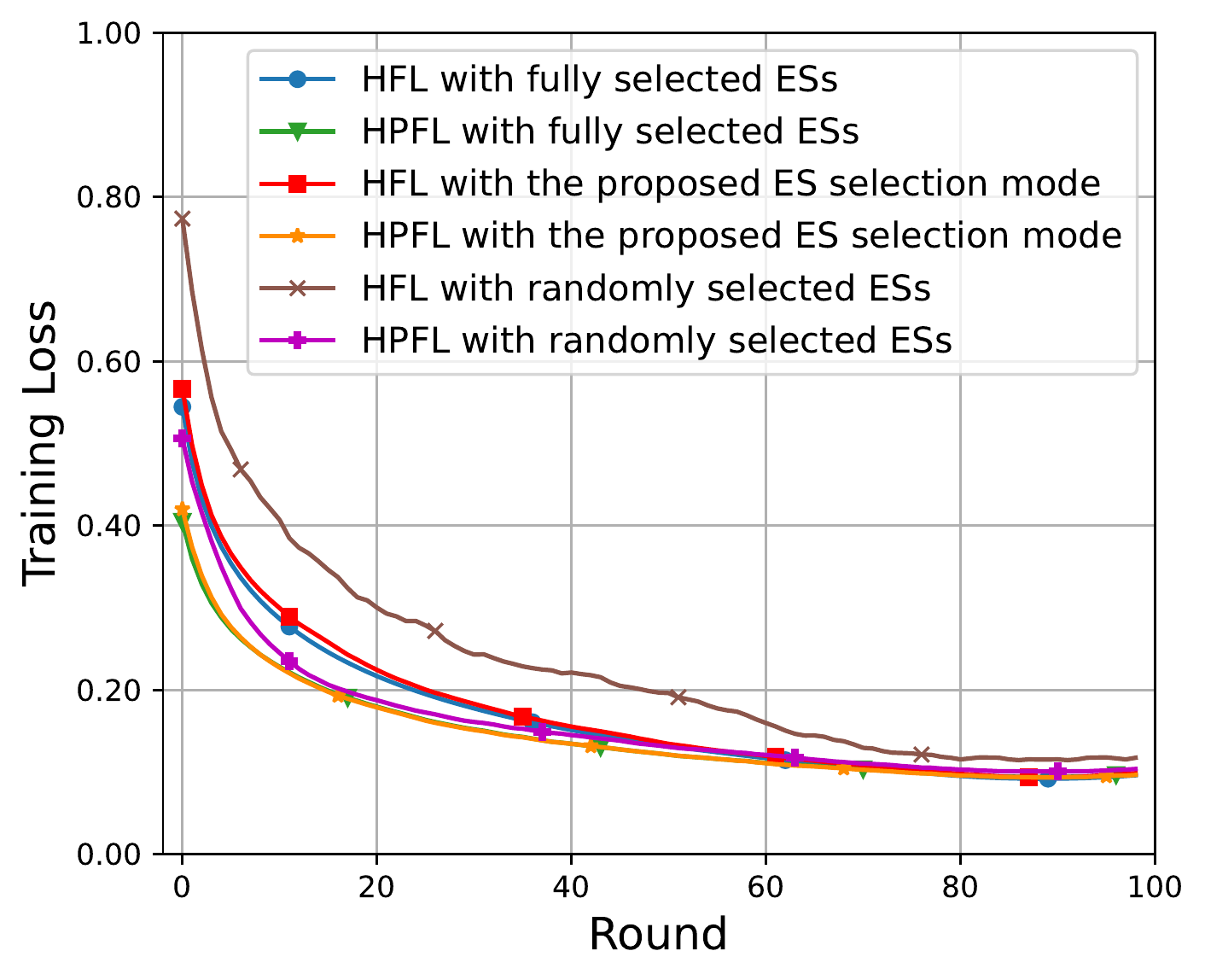}
      \label{fig_exp_1:subfig:a}}
  \subfloat[MNIST test accuracy]{
      \includegraphics[width=3in]{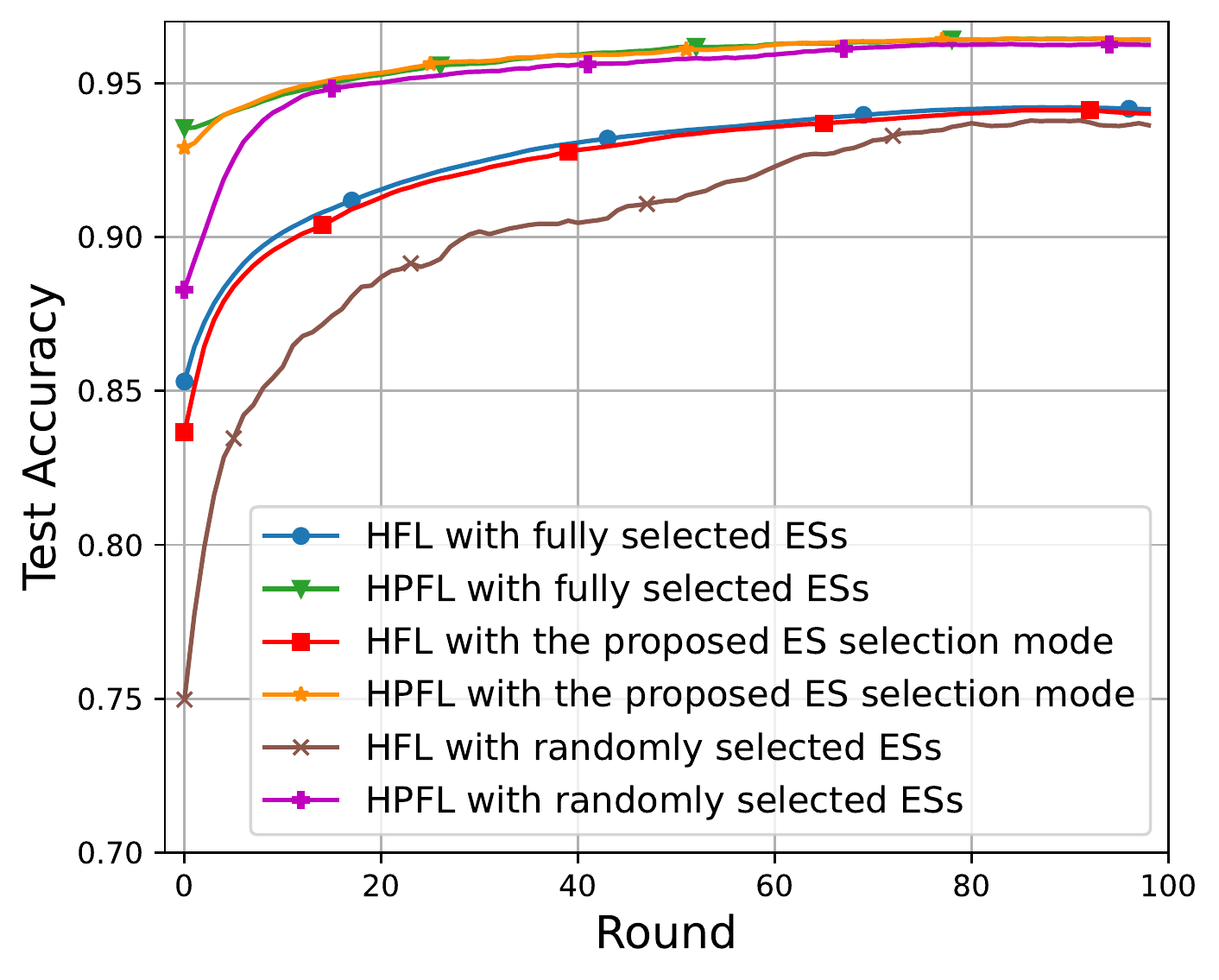}
      \label{fig_exp_1:subfig:b}}
  \caption{Convergence performance of HFL and HPFL with fully selected ESs, randomly selected ESs and the proposed ES selection mode using MNIST dataset. In this case, $\rho = 0.8$, and $l = 2$.}
  \label{fig:exp_1}
\end{figure*}

\begin{figure*}[!t]
  \centering
  \subfloat[CIFAR-10 training loss]{
      \includegraphics[width=3in]{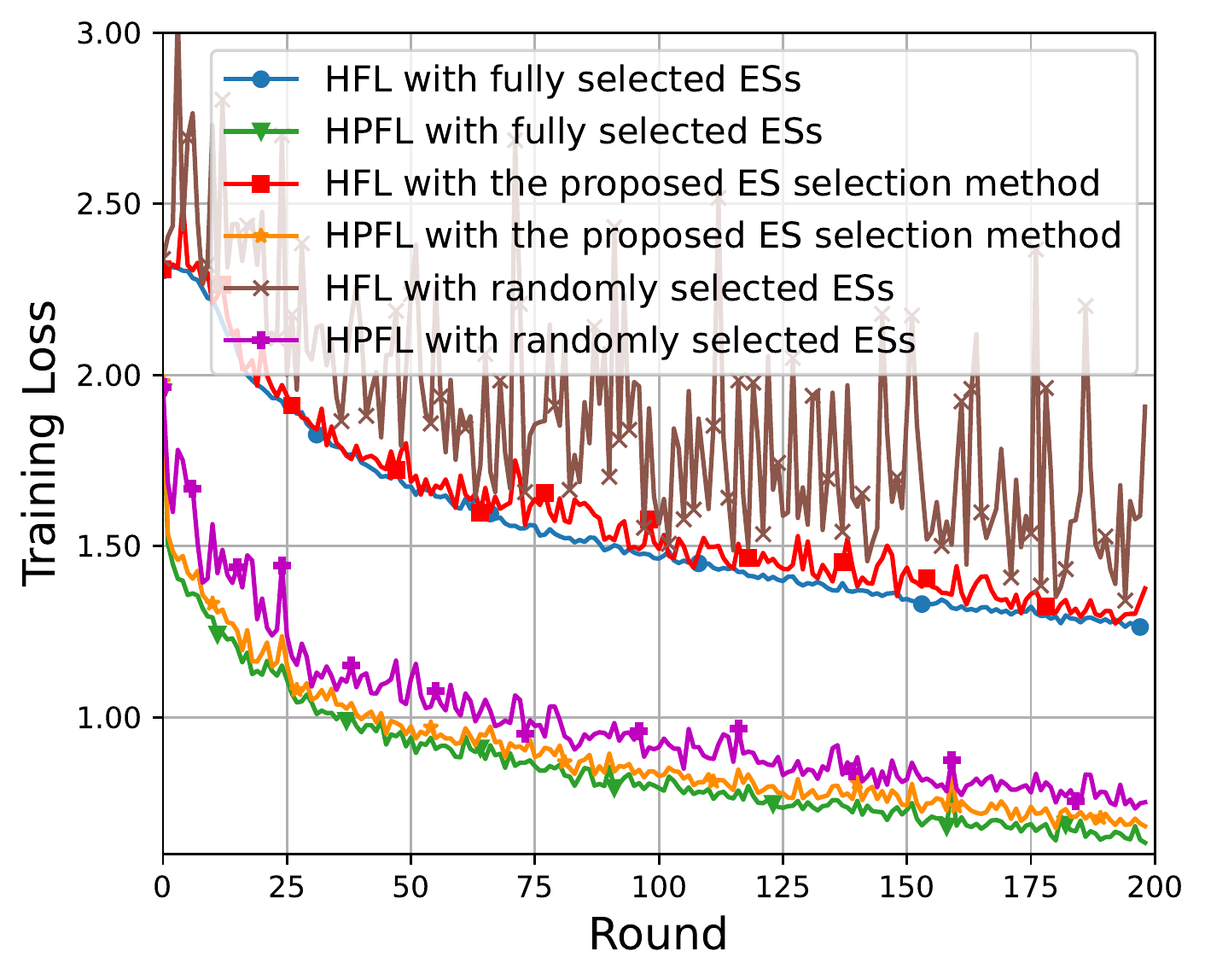}
      \label{fig_exp_2:subfig:a}}
  \subfloat[CIFAR-10 test accuracy]{
      \includegraphics[width=3in]{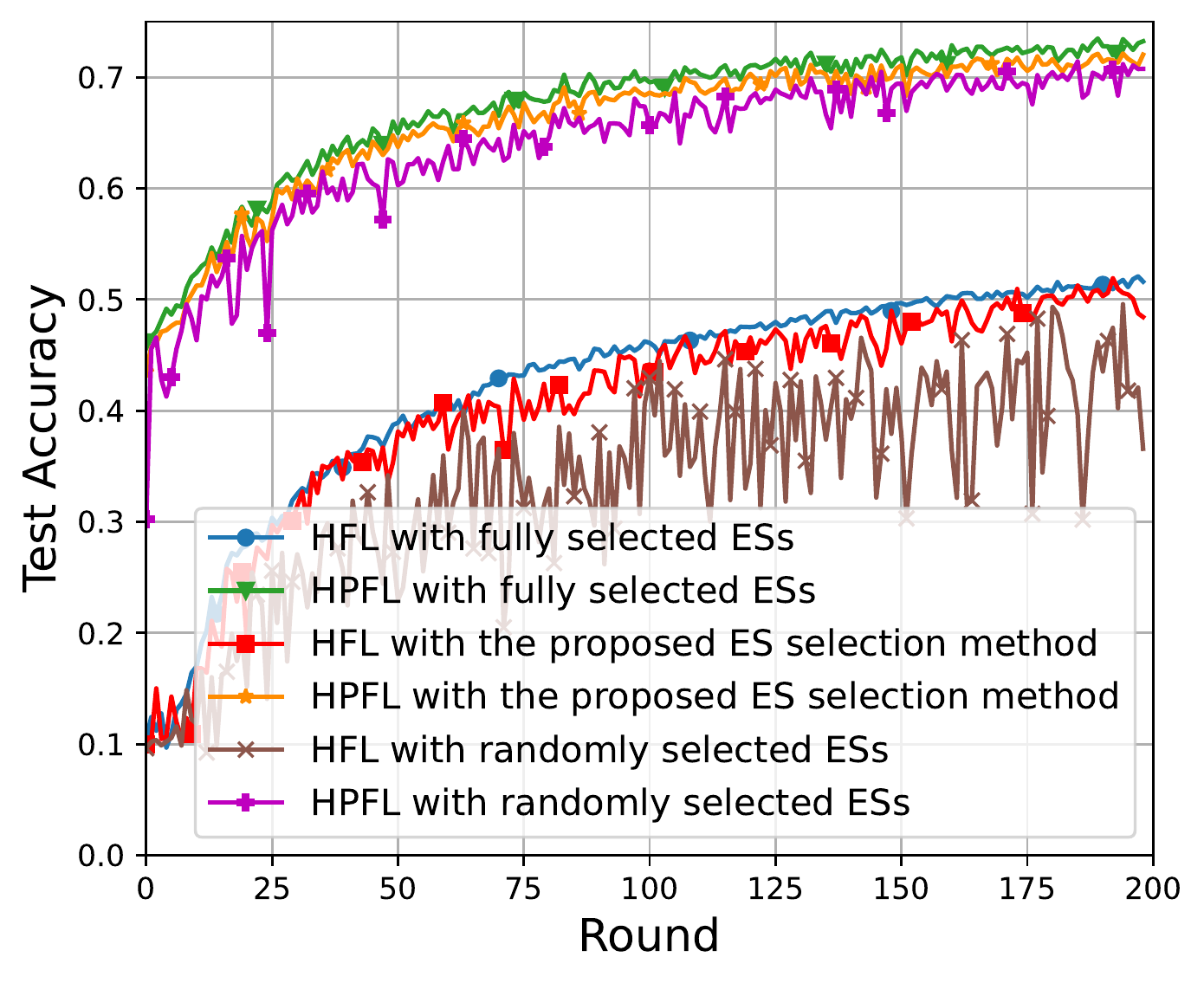}
      \label{fig_exp_2:subfig:b}}
  \caption{Convergence performance of HFL and HPFL with fully selected ESs, randomly selected ESs and the proposed ES selection mode using CIFAR-10 dataset. In this case, $\rho = 0.8$, and $l = 2$.}
  \label{fig:exp_2}
\end{figure*}

\subsubsection{Performance Metrics}

For the convergence performance, we measure the training loss as well as the test accuracy of the algorithms. The smaller the training loss is, the better the algorithm behaves.
On the contrary, the larger the test accuracy is, the better the algorithm behaves.
Meanwhile, for the implementation performance of HPFL over MEC networks, we measure HPFL's round latency and its total data importance collected in each round.
We also measure the runtime of implementing HPFL in MEC networks compared with intuitive algorithms with no needs to compute the ES scheduling policy and bandwidth allocation.

\begin{figure*}[!t]
  \centering
  \subfloat[MNIST training loss]{
      \includegraphics[width=2.8in]{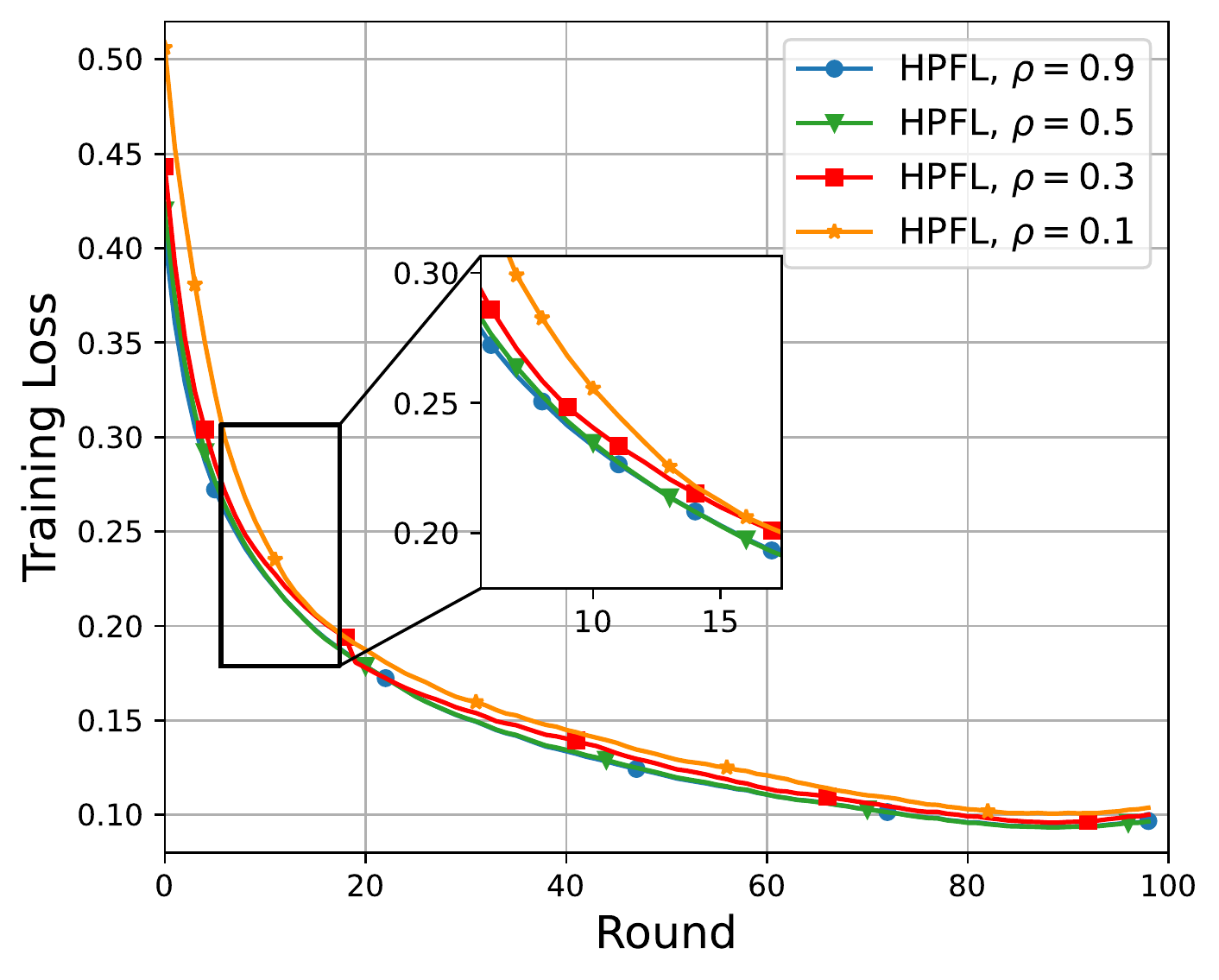}
      \label{fig_exp_3:subfig:a}}
  \subfloat[MNIST test accuracy]{
      \includegraphics[width=2.8in]{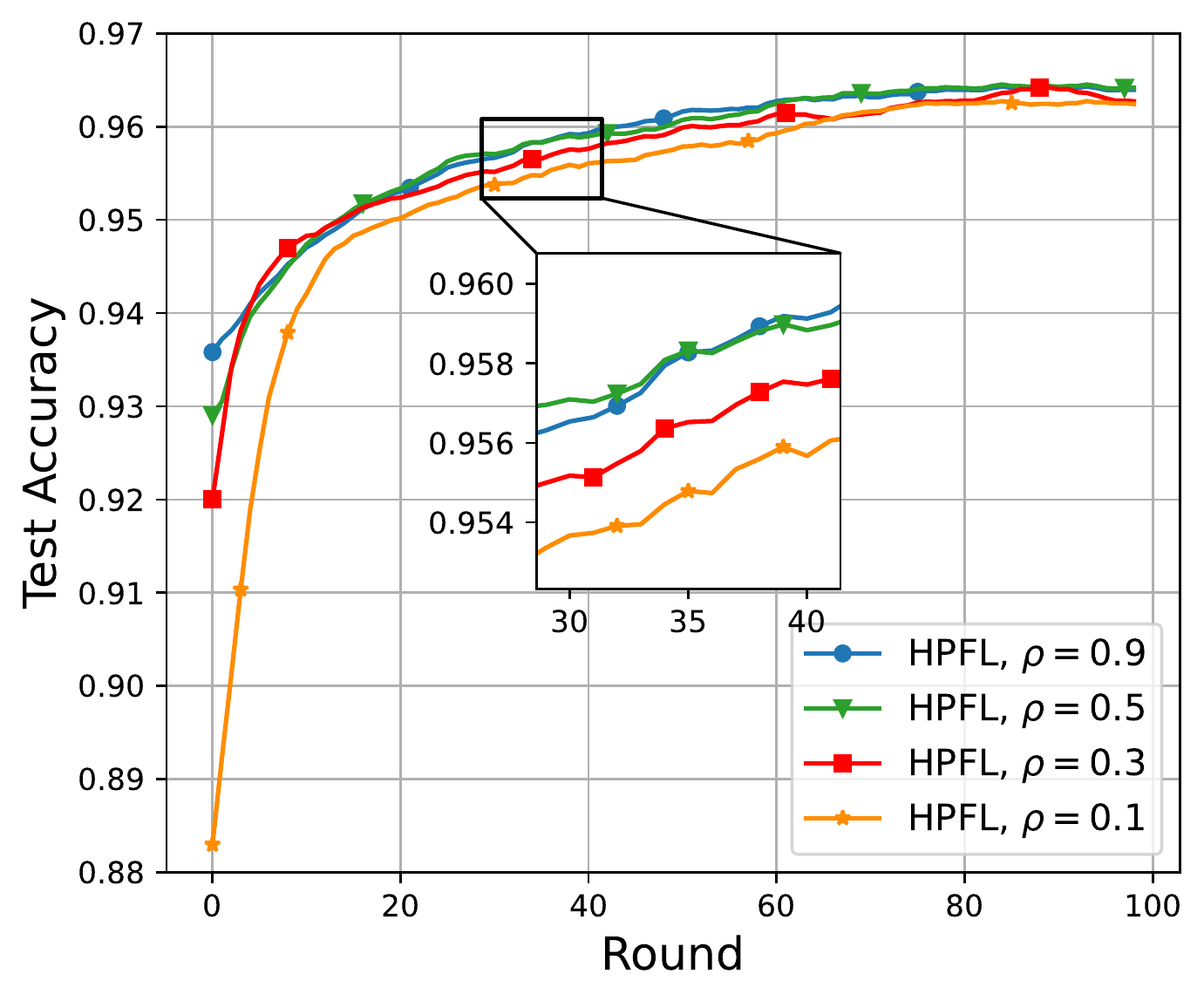}
      \label{fig_exp_3:subfig:b}}
  \caption{Convergence performance of our proposed HPFL using MNIST dataset. The weight factor $\rho$ is set to be $0.1, 0.3, 0.5, 0.9$.}
  \label{fig:exp_3}
\end{figure*}

\begin{figure*}[!t]
  \centering
  \subfloat[CIFAR-10 training loss]{
      \includegraphics[width=2.8in]{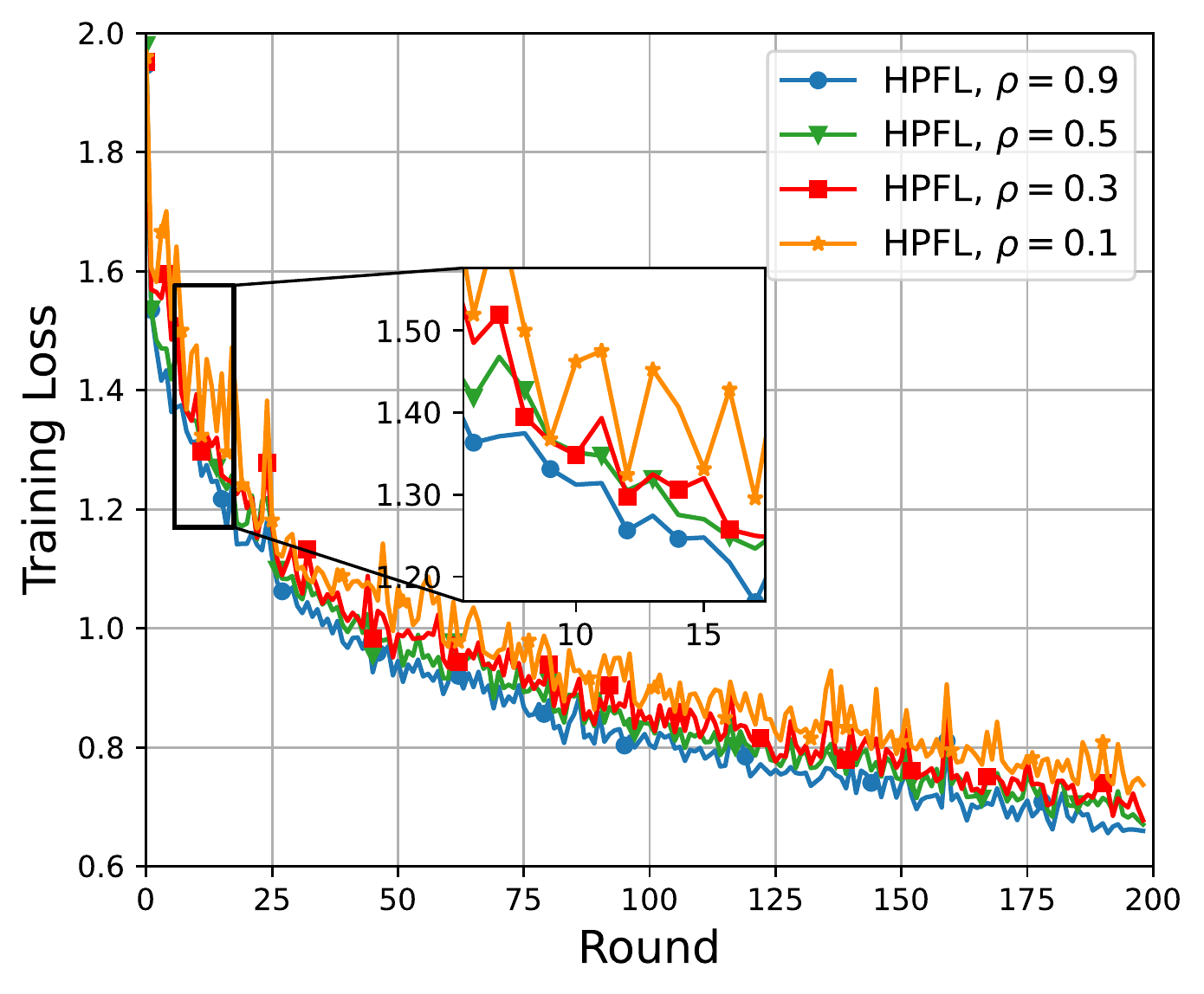}
      \label{fig_exp_4:subfig:a}}
  \subfloat[CIFAR-10 test accuracy]{
      \includegraphics[width=2.8in]{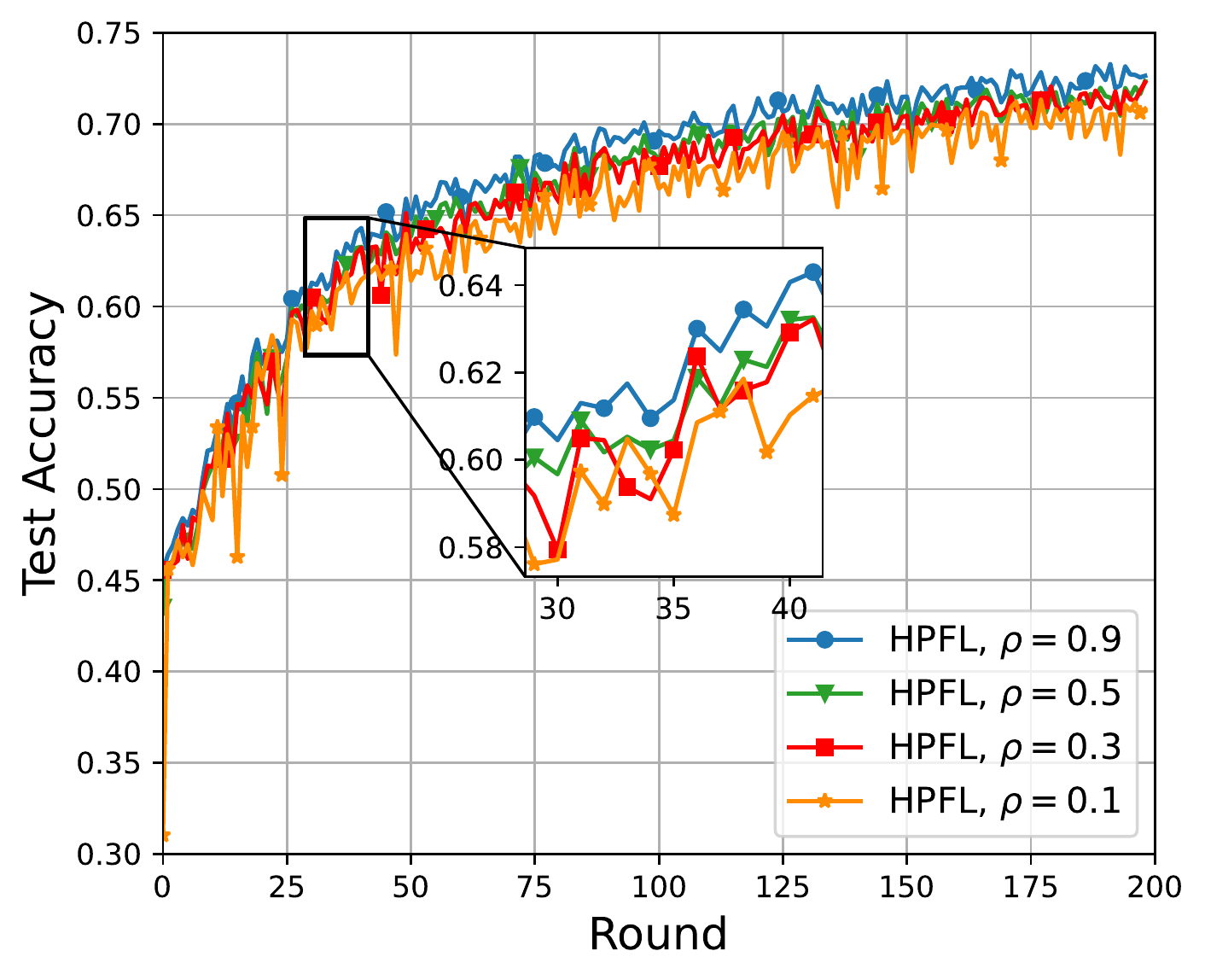}
      \label{fig_exp_4:subfig:b}}
  \caption{Convergence performance of our proposed HPFL using CIFAR-10 dataset. The weight factor $\rho$ is set to be $0.1, 0.3, 0.5, 0.9$.}
  \label{fig:exp_4}
\end{figure*}

\subsection{Convergence Performance} \label{sec:7.2}

We begin with evaluating the convergence performance of the six algorithms: HFL and HPFL with fully selected ESs, randomly selected ESs and the proposed ES selection mode, respectively.
Fig.~\ref{fig:exp_1} illustrates the convergence performance of the six algorithms under the MNIST dataset, while Fig.~\ref{fig:exp_2} shows the same convergence performance using the CIFAR-10 dataset.
%
%
From both figures, we observe that the algorithms with fully selected ESs always outperform the others.
This aligns with our intuition as all-in training provides information about all ESs and all UEs.
We also notice that the algorithms with randomly selected ESs behave the worst. This is also a reasonable result because ESs that are randomly selected can not always provide the most effective information for the global model training.
Meanwhile, all the HPFL algorithms outperform the HFL algorithms since the HPFL always behaves well in dealing with heterogeneous datasets. This result provides another proof of PFL's advantage in fast convergence, even though it is implemented using hierarchical aggregations.

For the MNIST dataset, it is a simple dataset that is easy to train. Therefore, if HPFL performs well, its behavior should be pretty close to the optimal one, where all ESs are selected in each communication round. The results shown in Fig.~\ref{fig:exp_1} just prove the above point, that the proposed ES scheduling scheme gives almost the same result as the fully selected ES scheme.
Moreover, as we can observe from Fig.~\ref{fig:exp_2} where the CIFAR-10 dataset is considered, the convergence performance of algorithms with fully selected ESs is more stable than that of algorithms with the proposed ES selection method and the randomly selected ESs, while the algorithms with randomly selected ESs behave the worst. This is a reasonable result, that an all-in UE selection mode contains data information from all UEs, whereas a randomly UE selection mode may leave out important data information. Besides, algorithms with the proposed ES selection method attempt to pick up the UEs with more data importance in each rounds, thereby leading to a convergence performance in between.

A detailed statistics breakdown is given in Fig.~\ref{fig:exp_3} and Fig.~\ref{fig:exp_4}. These two figures show the convergence performance of the proposed HPFL under different values of $\rho$, i.e., $\rho = 0.1, 0.3, 0.5, 0.9$. We observe that the higher the value of $\rho$ is, the better the algorithm's training loss and test accuracy are. We attribute this phenomenon to the impact of $\rho$, that when $\rho$ increases, HPFL is inclined to choose ESs with higher data importance. Therefore, the HPFL's convergence behavior under higher values of $\rho$ would be better.

\begin{figure*}[!t]
  \centering
  \subfloat[MNIST dataset]{
      \includegraphics[width=3in]{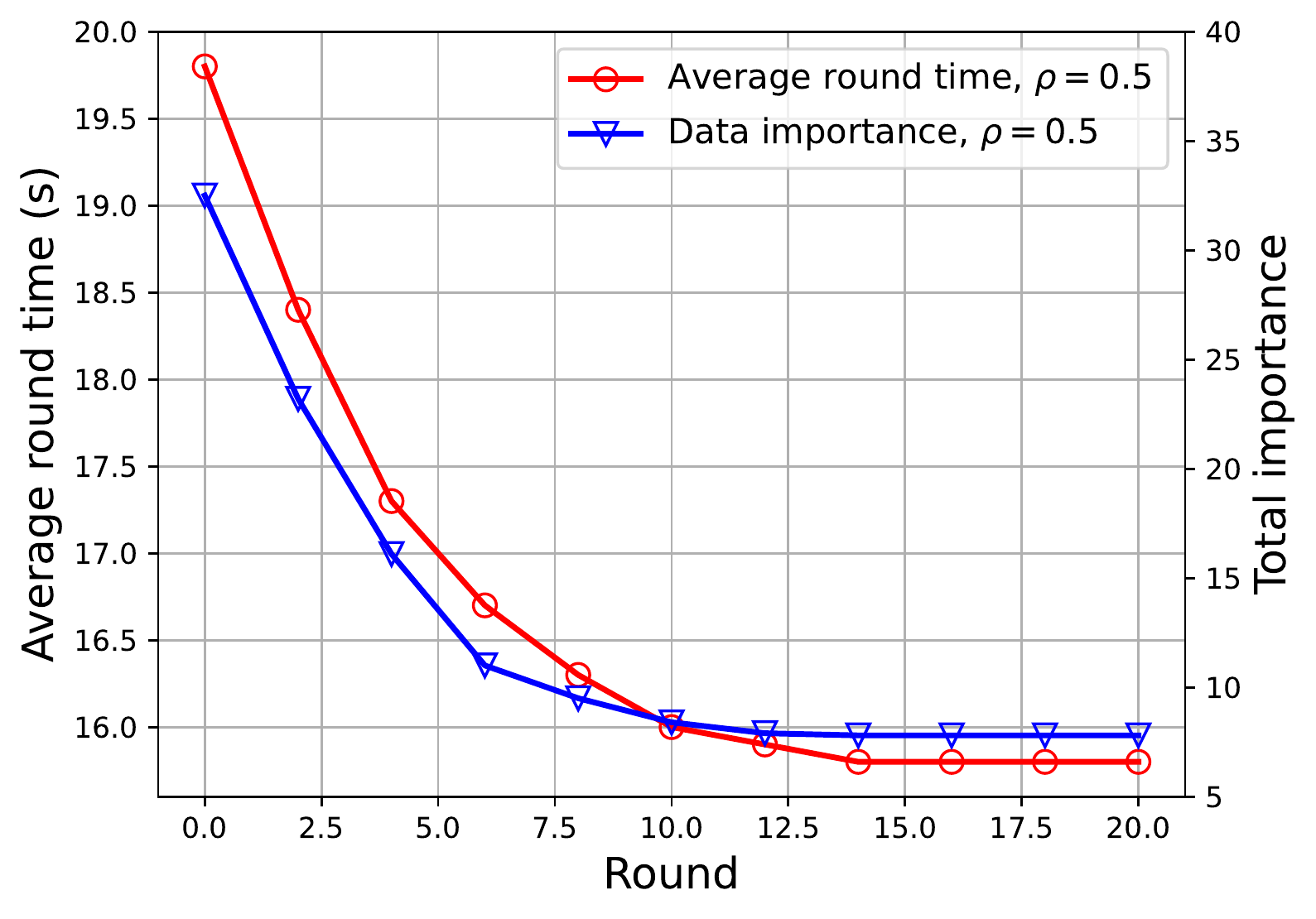}
      \label{fig_exp_5:subfig:a}}
  \subfloat[CIFAR-10 dataset]{
      \includegraphics[width=3in]{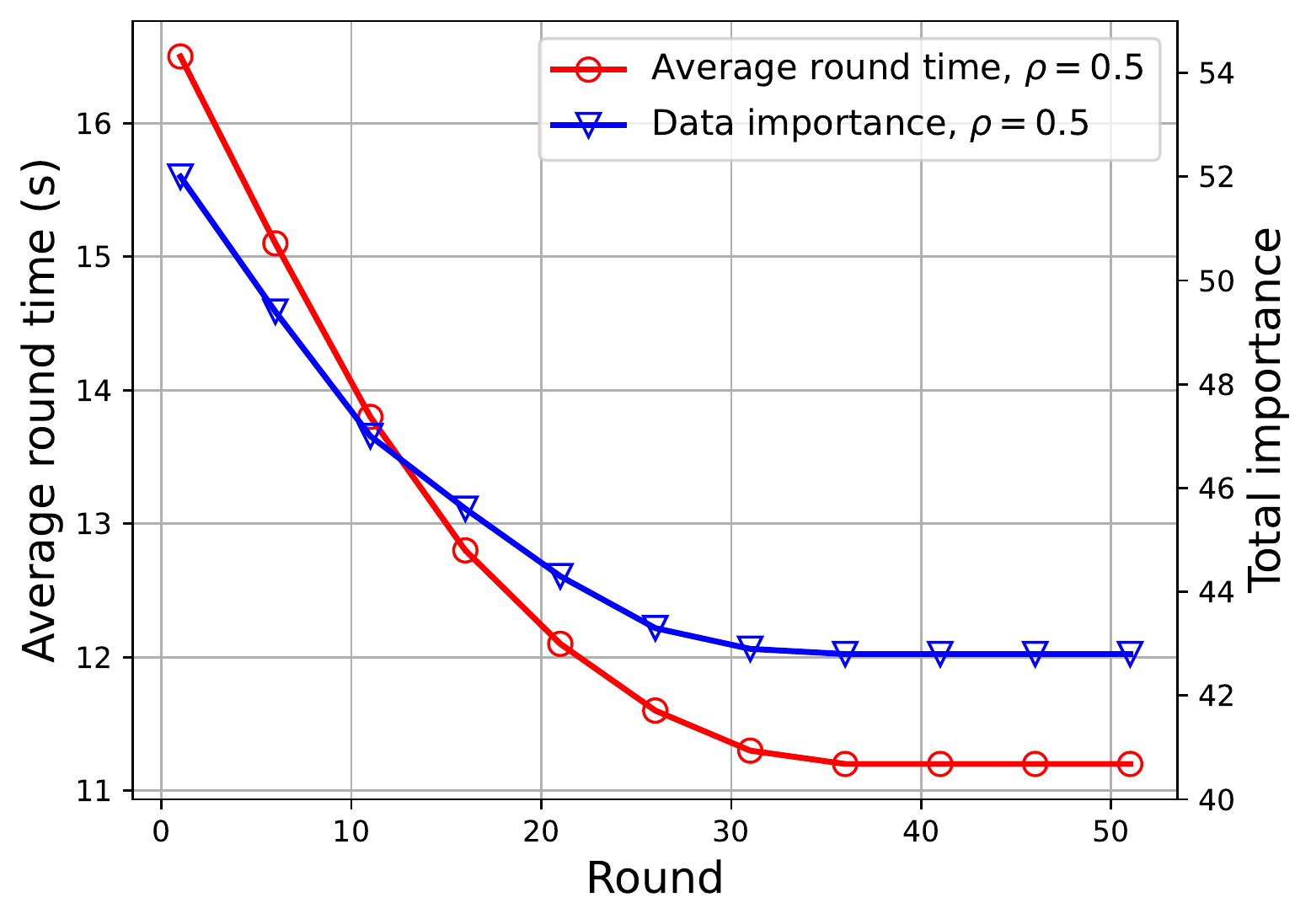}
      \label{fig_exp_5:subfig:b}}
  \caption{Average round latency and round data importance performance vs. the number of rounds}.
  \label{fig:exp_5}
\end{figure*}

\begin{figure*}[!t]
  \centering
  \subfloat[Average round latency]{
      \includegraphics[width=2.8in]{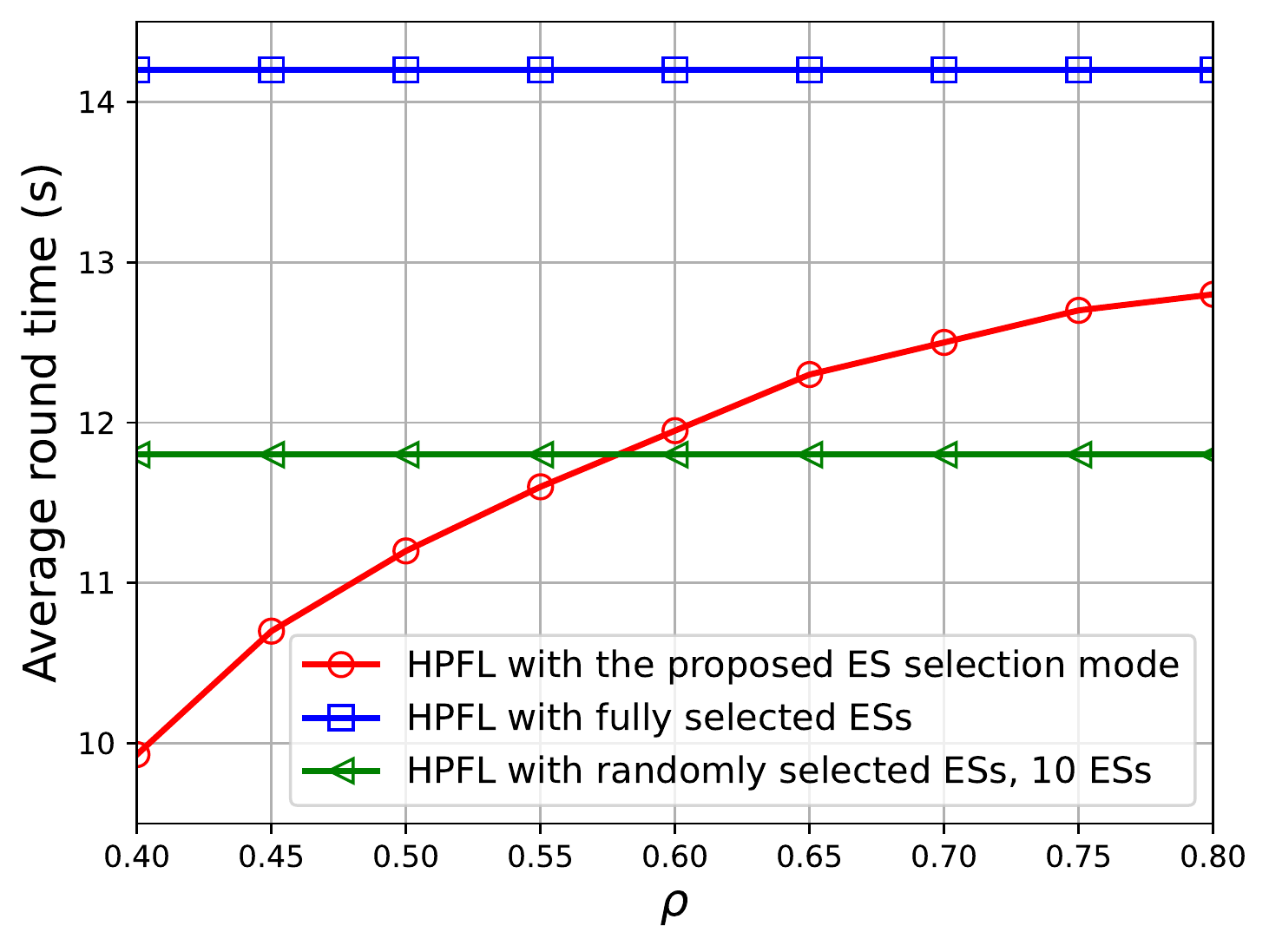}
      \label{fig_exp_6:subfig:a}}
  \subfloat[Total data importance]{
      \includegraphics[width=2.8in]{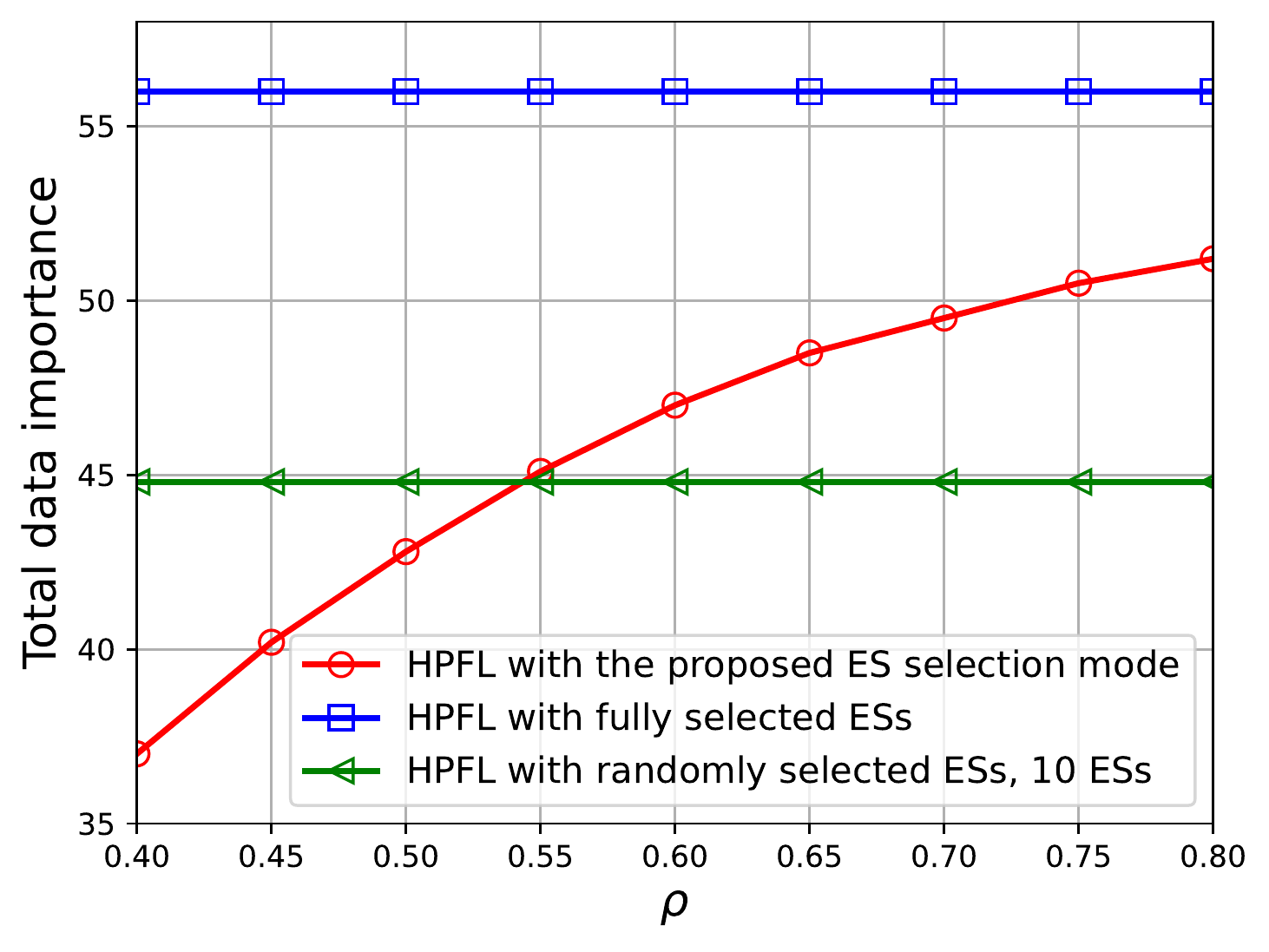}
      \label{fig_exp_6:subfig:b}}
  \caption{Average round latency and round data importance performance with MNIST dataset.}
  \label{fig:exp_6}
\end{figure*}

\subsection{Average Round Latency vs Total Data Importance} \label{sec:7.3}

In this subsection, we examine the HPFL average round latency and total data importance captured in each round with respect to the number of communication rounds and the weight factor $\rho$, respectively.

\subsubsection{Effect of the communication round}

From Fig.~\ref{fig:exp_5}, we observe that both average round latency and total data importance  decrease as the number of rounds increases.
This is because the objective of HPFL is to minimize the round latency and the round training loss.
Both optimal round latency and optimal total data importance captured in each round are bound to continuously decrease as the round number increases.
Such a decline slows down as HPFL converges, as it is shown in Fig.~\ref{fig:exp_5}.

\subsubsection{Effect of the weight factor $\rho$}

Next, we investigate the effect of $\rho$ on the average round time and the total data importance captured in each round.
Due to space concerns, we only used the MNIST dataset to do the experiments. The value of $\rho$ changes from $0.4$ to $0.8$ under the step of $0.05$, and the results are measured when the HPFL algorithms converge, which are shown in Fig.~\ref{fig:exp_6}.
We observe that for HPFL algorithms with fully selected and randomly selected UEs, both metrics remain unchanged under various weight factors.
However, as for the HPFL algorithm with the proposed ES selected mode, the average round latency and the total data importance captured in each round increase as the value of $\rho$ increases.
This is a reasonable result because, as the weight factor increases, it is desired for the scheduler to choose ESs with larger total data importance. Therefore, as we can see in Fig.~\ref{fig_exp_6:subfig:b}, the larger $\rho$ is, the more total data importance gathered in each round. At the same time, as the HPFL's objective function indicates, more selected data importance means larger round latency, and the result shown in Fig.~\ref{fig:exp_6} exactly verifies this point of view.

\subsubsection{Effect of the heterogenous factor $l$}

\begin{figure*}[!t]
  \centering
  \subfloat[MNIST training loss]{
      \includegraphics[width=2.8in]{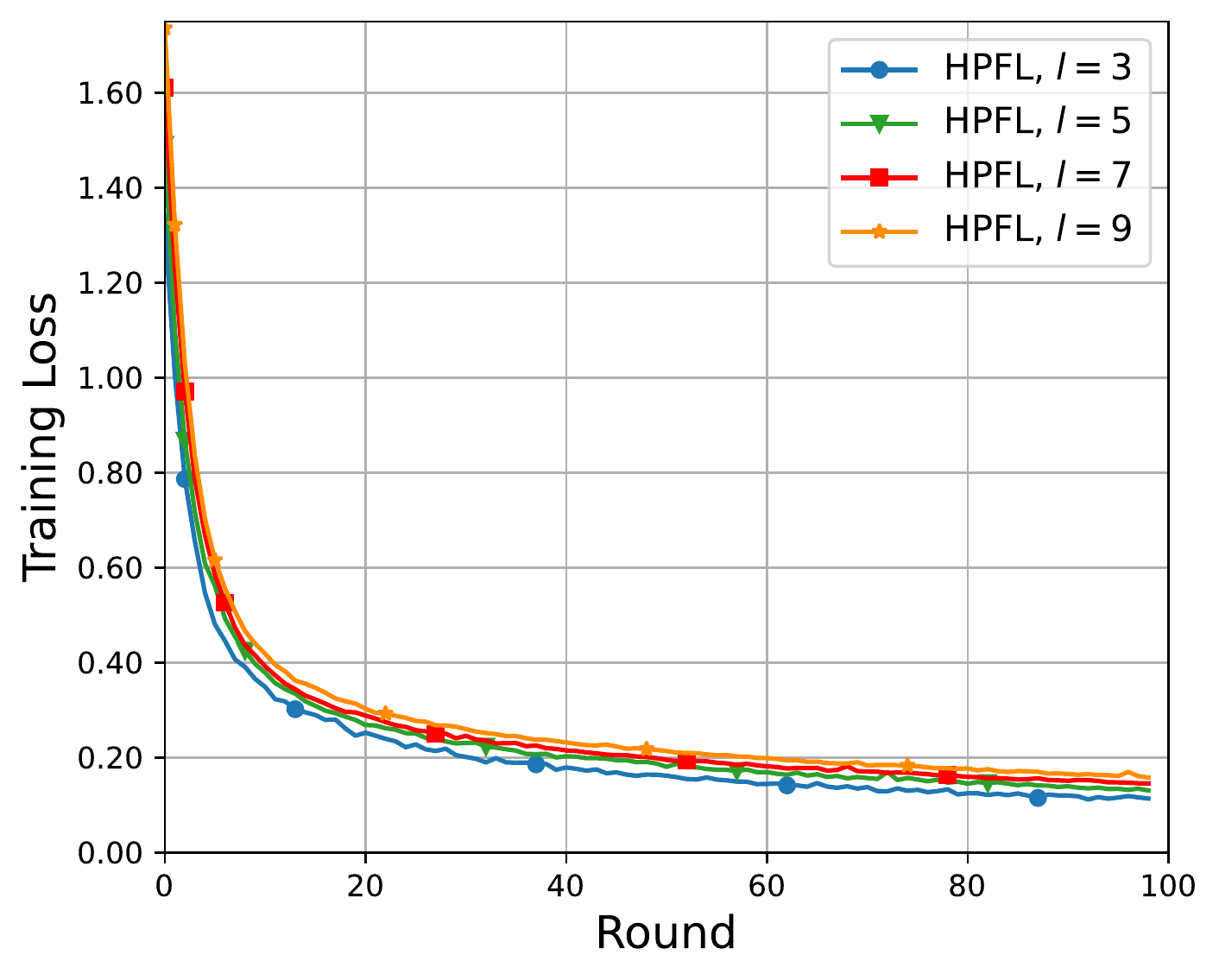}
      \label{fig_exp_8:subfig:a}}
  \subfloat[MNIST test accuracy]{
      \includegraphics[width=2.8in]{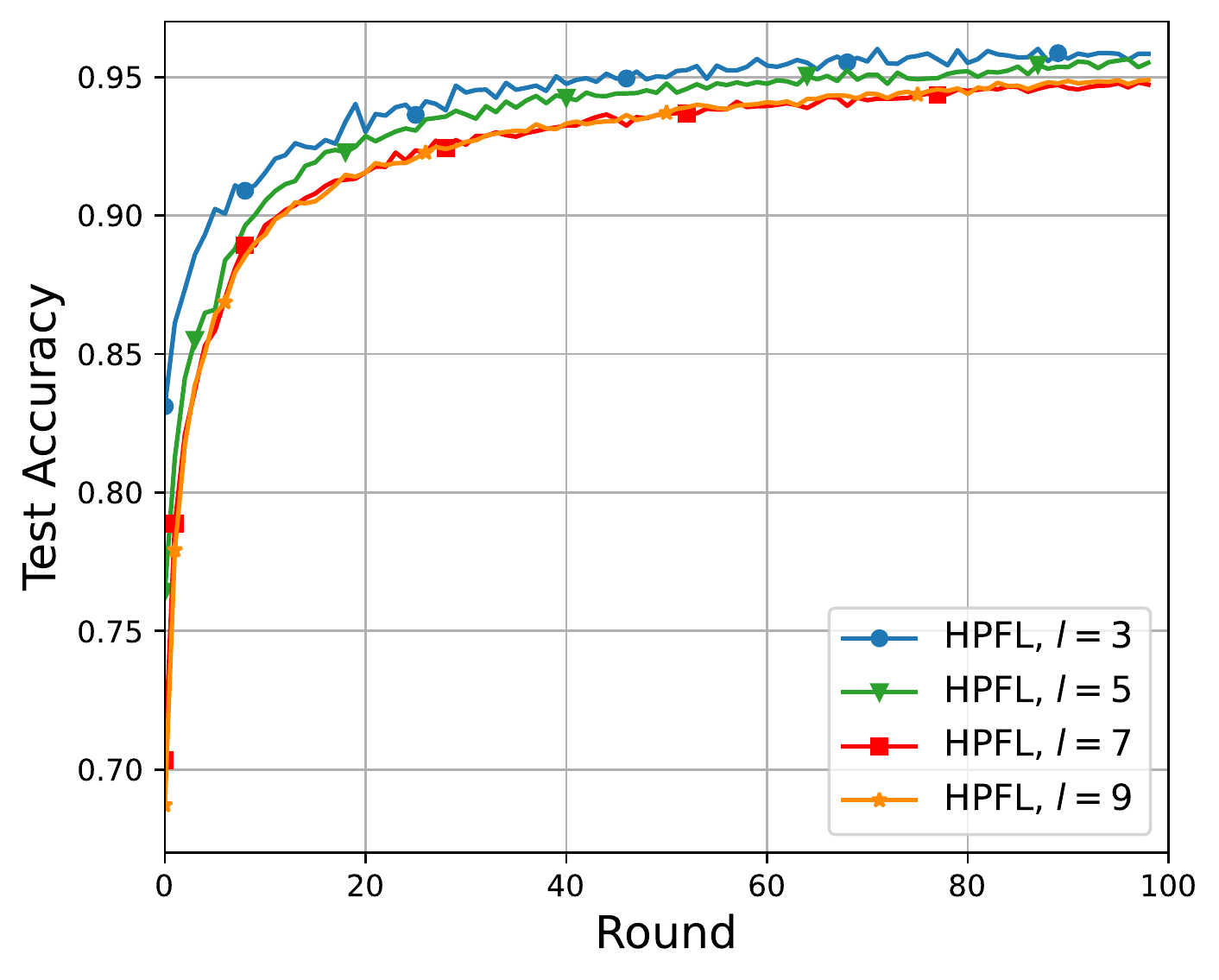}
      \label{fig_exp_8:subfig:b}}
  \caption{Convergence performance of our proposed HPFL using MNIST dataset. The heterogenous level $l$ is set to be $3, 5, 7, 9$.}
  \label{fig:exp_8}
\end{figure*}

\begin{figure*}[!t]
  \centering
  \subfloat[CIFAR-10 training loss]{
      \includegraphics[width=2.8in]{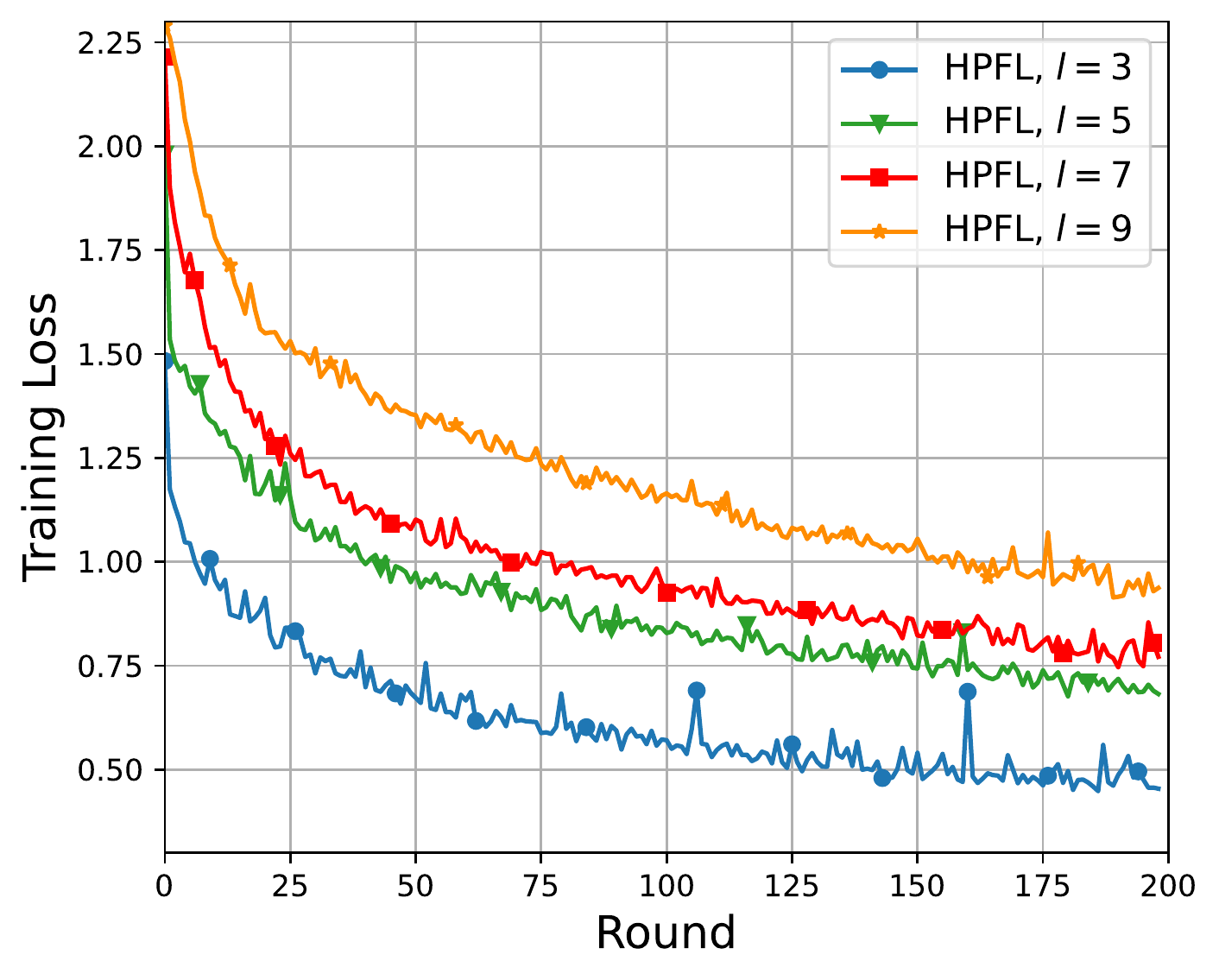}
      \label{fig_exp_9:subfig:a}}
  \subfloat[CIFAR-10 test accuracy]{
      \includegraphics[width=2.8in]{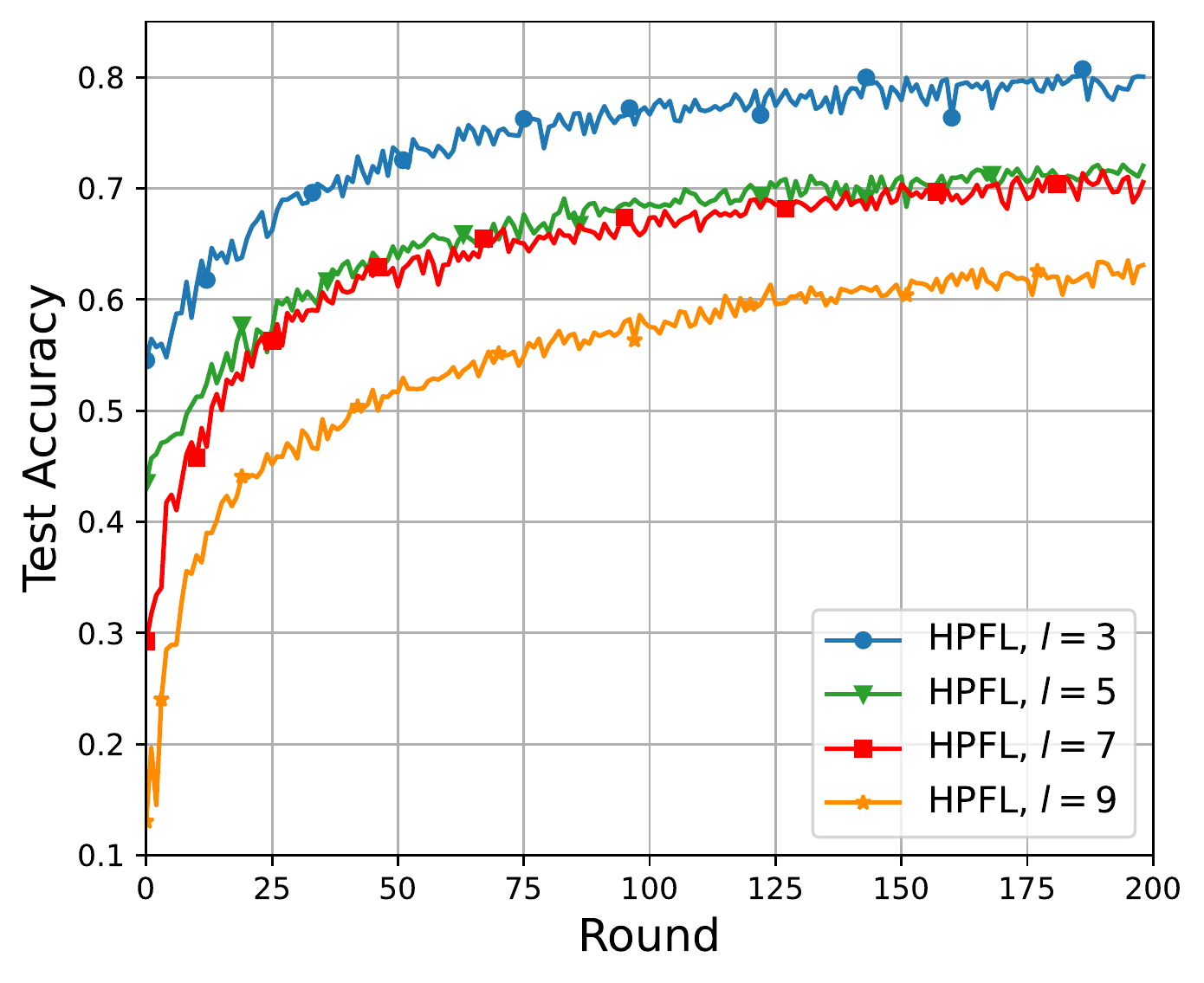}
      \label{fig_exp_9:subfig:b}}
  \caption{Convergence performance of our proposed HPFL using CIFAR-10 dataset. The heterogenous level $l$ is set to be $3, 5, 7, 9$.}
  \label{fig:exp_9}
\end{figure*}

Now we come to explore the effect of the heterogenous level $l$ on the learning performance of HPFL. The results are shown in Fig.~\ref{fig:exp_8} and \ref{fig:exp_9}. From both datasets we observe that the larger the heterogenous level $l$ is, the poorer convergence performances are. This phenomenon is more pronounced in the CIFAR-10 dataset than that in the MNIST dataset. This result shows that the more complicated the dataset is, the difference between UEs with different levels of heterogeneity is more obvious.

\subsection{Algorithm Runtime Performance} \label{sec:7.4}

\begin{figure*}[!t]
  \centering
  \subfloat[Both datasets]{
      \includegraphics[width=2.7in]{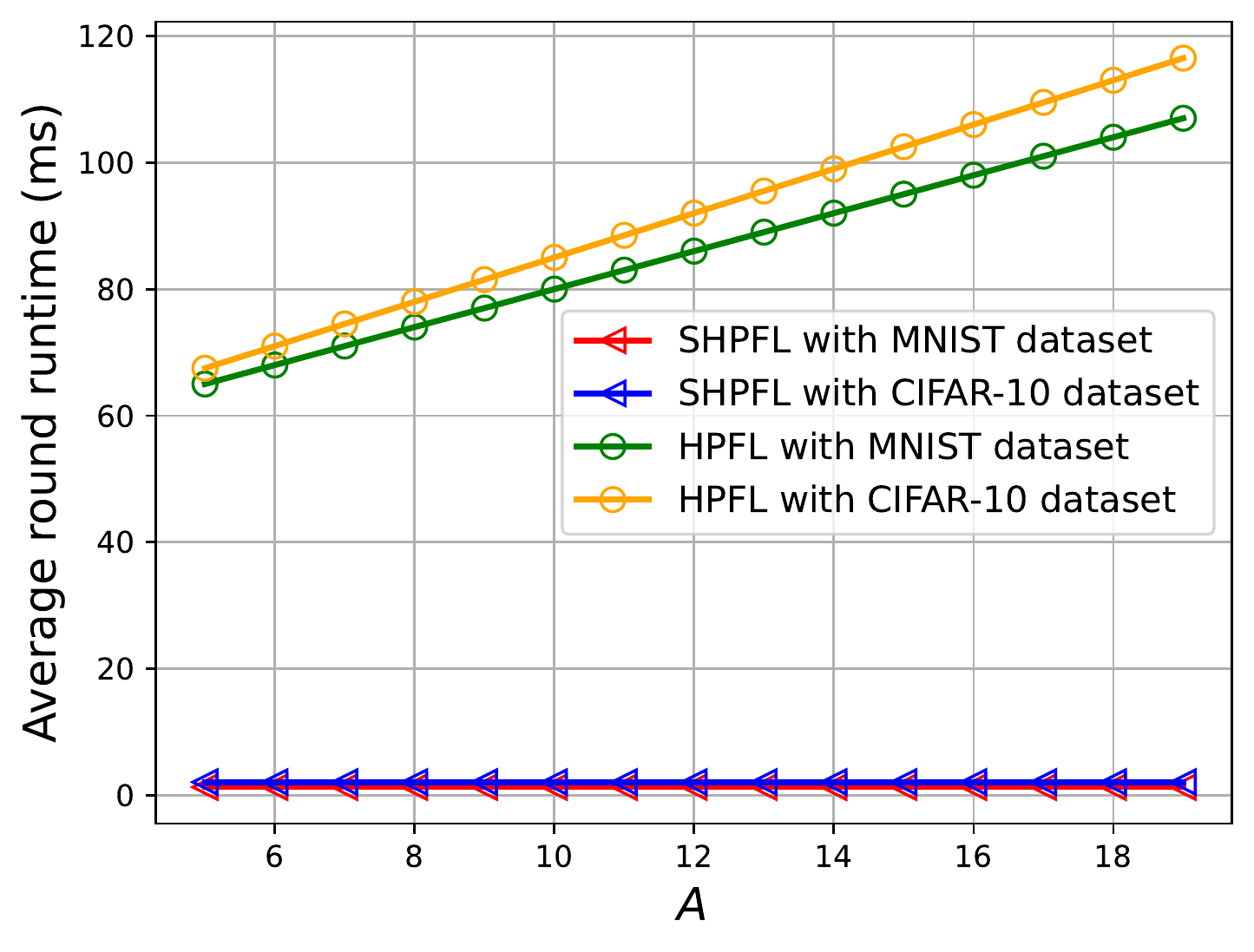}
      \label{fig_exp_7:subfig:a}}
  \subfloat[CIFAR-10 dataset]{
      \includegraphics[width=3.05in]{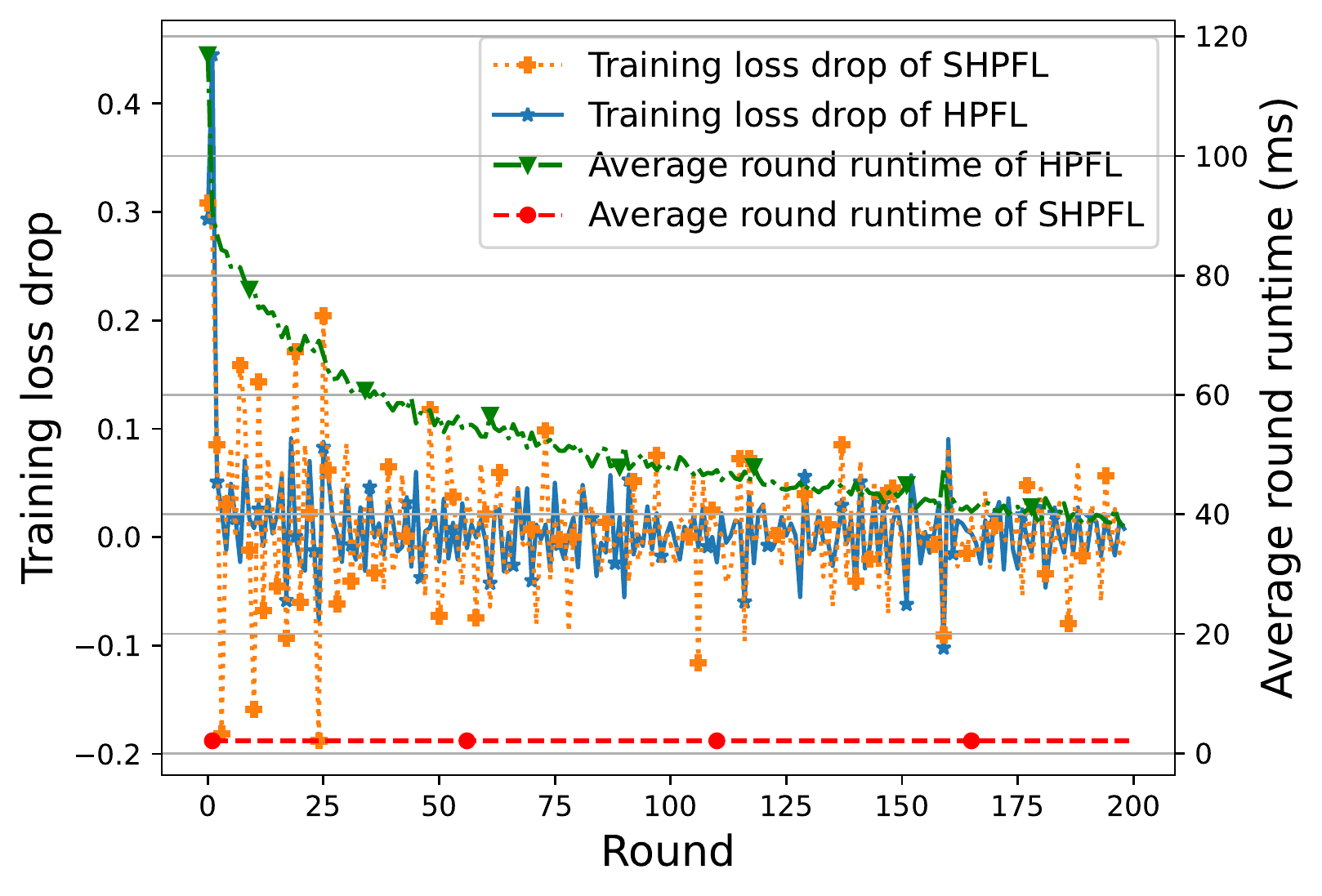}
      \label{fig_exp_7:subfig:b}}
  \caption{Average round runtime performance where (a) uses the MNIST and CIFAR-10 datasets and (b) uses only the CIFAR-10 dataset. In this case, $\rho = 0.8$, and $l=2$.}
  \label{fig:exp_7}
\end{figure*}

Finally, we check the price HPFL pays for its gains in saving round latencies and its advantages in selecting ESs with more data importance.
We compare the runtime of HPFL and a simple hierarchical PFL algorithm, where ESs are also scheduled semi-asynchronously but randomly selected, whereas the bandwidth is equally divided, namely, $b_t^{k,1} = b_t^{k,2} = \dots = b_t^{k, n_k}$, and $b_t^{k,0} = b_t^{j,0}$ $(k\neq j, k,j\in\mathcal{A}_t)$.
We name this simple hierarchical PFL algorithm as SHPFL. Meanwhile, the runtime is measured as the average duration when \texttt{BandwidthAllocation} and \texttt{ESScheduling} algorithms are triggered in each round. There is no need for SHPFL to compute the ES scheduling policy and the bandwidth allocation problem. Therefore, as shown in Fig.~\ref{fig_exp_7:subfig:a}, as the number of selected ESs increases, the average runtime of SHPFL in each round keeps constant, while that of HPFL grows linearly.
Moreover, the average runtime in each round of HPFL algorithms is at least $60 \times$ larger than that of SHPFL. This indicates that to save more round latency and capture more data information, HPFL sacrifices its runtime for better convergence and implementation performance. Nevertheless, the consumed runtime (in tens of milliseconds) of our proposed HPFL is acceptable since it is much smaller than the average round time (in tens of seconds).

Fig.~\ref{fig_exp_7:subfig:b} shows the tradeoff between average round runtime and the training loss drop along with the communication rounds. It is obvious that for the proposed HPFL algorithm, the average round runtime and the training loss drop are positively correlated. This is a reasonable result, since as the communication rounds increases, the training loss drop decreases. According to the ES scheduling algorithm, the number of selected ESs in each round decreases with the decreasing of the training loss drop, thereby leading to the reduction of average round runtime. However, for the SHPFL algorithm, the average round runtime does not change and has no relationship witht eh training loss drop. This is also a reasonable result, since there is no need for SHPFL to compute the ES scheduling policy and the bandwidth allocation problem.

\section{Conclusions} \label{sec:8}

In this paper, we have studied the hierarchical PFL mechanism for massive MEC networks.
We have proposed HPFL, a hierarchical PFL algorithm that adapts quickly to individual UEs with hierarchical aggregations.
HPFL is implemented in a three-layer MEC network, where ESs aggregates the edge models synchronously, and the CS fuses the global model update in a semi-asynchronous manner.
The semi-asynchronous aggregation at the CS brings a tradeoff between the round training loss and the round latency. HPFL combines the objectives of training loss maximization and round latency minimization while determining the ES scheduling policy and bandwidth allocation.
This optimization problem is solved by firstly analyzing the convergence of HPFL, thus transforming the problem into a tractable one. Then, we decoupled the problem into an ES scheduling problem and a bandwidth allocation problem, and solved them separately. The extensive experimental results demonstrated that not only HPFL is convergent, but provides a tradeoff in maximizing the round training loss and minimizing the round latency.

\section*{Appendix}

\noindent\textbf{Proof of Theorem 1}

\vspace{0.1cm}
To begin with, let
\begin{align}
  & \sum_{k=1}^{K}\sum_{i=1}^{n_k} \frac{\pi_t^k}{A n_k} \nabla F_{k,i}(\mathbf{w}_{t-\tau_t^k}) \nonumber \\
  = & X+Y+\sum_{k=1}^{K}\sum_{i=1}^{n_k} \frac{\pi_t^k}{A n_k} \nabla F(\mathbf{w}_{t-\tau_t^k}),
\end{align}
where
\begin{align}
  X = & \sum_{k=1}^{K}\sum_{i=1}^{n_k} \frac{\pi_t^k}{A n_k}\left(\nabla F_{k,i}(\mathbf{w}_{t-\tau_t^k}) - \nabla F_k(\mathbf{w}_{t-\tau_t^k})\right), \nonumber \\
  Y = & \sum_{k=1}^{K}\sum_{i=1}^{n_k} \frac{\pi_t^k}{A n_k}\left( \nabla F_k(\mathbf{w}_{t-\tau_t^k}) - \nabla F(\mathbf{w}_{t-\tau_t^k})\right).
\end{align}
We next bound the moments of $X$ and $Y$. According to the Cauchy-Schwarz inequality
\begin{equation}
  \left\|\sum_{k=1}^{K} a_k b_k \right\|^2 \leq \left(\sum_{k=1}^{K} \|a_k\|^2\right) \left( \sum_{k=1}^{K} \|b_k\|^2 \right).
\end{equation}
as for $X$, if we consider the inequality with $a_k = \frac{\pi_t^k}{A\sqrt{n_k}}$ and $b_k = \frac{1}{\sqrt{n_k}} (\nabla F_{k,i}(\mathbf{w}_{t-\tau_t^k}) - \nabla F_k(\mathbf{w}_{t-\tau_t^k}))$, then we have
\begin{align}
  \|X\|^2 \leq & \sum_{k=1}^{K} \sum_{i=1}^{n_k} \left(\frac{\pi_t^k}{A\sqrt{n_k}}\right)^2 \nonumber \\
  & \sum_{k=1}^{K} \sum_{i=1}^{n_k} \frac{1}{n_k} \left\|\nabla F_{k,i}(\mathbf{w}_{t-\tau_t^k}) - \nabla F_k(\mathbf{w}_{t-\tau_t^k})\right\|^2\nonumber \\
  = & \frac{K \gamma_F^2}{A}.
\end{align}
Likewise, as for $Y$, if we consider the Cauchy-Schwarz inequality with $b_k = \frac{1}{\sqrt{n_k}} (\nabla F_{k}(\mathbf{w}_{t-\tau_t^k}) - \nabla F(\mathbf{w}_{t-\tau_t^k}))$ and $a_k = \frac{\pi_t^k}{A\sqrt{n_k}}$, then we have
\begin{equation}
  \|Y\|^2 \leq \frac{K \gamma_F^2}{A}.
\end{equation}
Meanwhile, Lemma~\ref{lem:1} indicates an inequation such that $F(\mathbf{w}_{t+1}) - F(\mathbf{w}_t) \leq \langle \nabla F(\mathbf{w}_t), \mathbf{w}_{t+1} - \mathbf{w}_t\rangle + \frac{L_F}{2} \| \mathbf{w}_{t+1} - \mathbf{w}_t \|^2$. Based on this inequation, if $\beta = \frac{1}{L_F}$, we have

\begin{align} \label{equ:convergence_rate}
 & F(\mathbf{w}_{t+1})- F(\mathbf{w}_t) \nonumber \\
 \leq & \langle\nabla F(\mathbf{w}_t), \mathbf{w}_{t+1} - \mathbf{w}_t \rangle + \frac{L_F}{2}\|\mathbf{w}_{t+1} - \mathbf{w}_t\|^2 \nonumber \\
 = & - \beta \nabla F(\mathbf{w}_t) \sum_{k\in\mathcal{K}} \sum_{i=1}^{n_k} \frac{\pi_t^k}{A n_k} \nabla F_{k,i}(\mathbf{w}_{t-\tau_t^k}) \nonumber \\
 & + \frac{L_F \beta^2}{2} \left \|\sum_{k\in\mathcal{K}}\sum_{i=1}^{n_k} \frac{\pi_t^k}{A n_k} \nabla F_{k,i}(\mathbf{w}_{t-\tau_t^k})\right\|^2 \nonumber \\
 = & -\beta \| \nabla F(\mathbf{w}_t)\|\left\| X + Y + \sum_{k=1}^k \sum_{i=1}^{n_k}\frac{\pi_t^k}{A n_k} \nabla F(\mathbf{w}_{t-\tau_t^k})\right\| \nonumber \\
 & + \frac{L_F \beta^2}{2}\left\|X+Y+ \sum_{k=1}^k \sum_{i=1}^{n_k} \frac{\pi_t^k}{A n_k} \nabla F(\mathbf{w}_{t-\tau_t^k}) \right\|^2 \nonumber \\
 \leq & \frac{5\beta}{2}\|X+Y\|^2 + \frac{\beta}{2}\left\|\sum_{k=1}^{K} \sum_{i=1}^{n_k} \nabla F(\mathbf{w}_{t-\tau_t^k})\right\|^2
  \nonumber \\
 & - \beta \| \nabla F(\mathbf{w}_t)\| \left\|\sum_{k=1}^{K} \sum_{i=1}^{n_k} \frac{\pi_t^k}{A n_k}\nabla F(\mathbf{w}_{t-\tau_t^k})\right\| \nonumber \\
 & + 2\beta \left \|\sum_{k=1}^{K} \sum_{i=1}^{n_k} \frac{\pi_t^k}{A n_k} \left(\nabla F(\mathbf{w}_t) -\nabla F(\mathbf{w}_{t-\tau_t^k})\right)\right\|^2 \nonumber \\
 \leq & \frac{5\beta}{2}\|X+Y\|^2 - \frac{\beta}{2} \nabla^2 F(\mathbf{w}_t)\nonumber \\
 & + \frac{5\beta}{2} \left\| \sum_{k=1}^{K} \frac{\pi_t^k}{A} \left( \nabla F(\mathbf{w}_t) -\nabla F(\mathbf{w}_{t-\tau_t^k}) \right) \right\|^2 \nonumber \\
 \leq & \frac{10 \beta K\gamma_F^2}{A} - \frac{\beta}{2} \nabla^2 F(\mathbf{w}_t)  \nonumber \\
 & + \frac{5\beta}{2} \left(\sum_{k=1}^{K} \frac{\pi_t^k}{A}\right) \left(\sum_{k=1}^{K} \frac{\pi_t^k}{A} \left\| \nabla F(\mathbf{w}_t) -\nabla F(\mathbf{w}_{t-\tau_t^k}) \right\|^2\right) \nonumber \\
 = & \frac{10 \beta K\gamma_F^2}{A} - \frac{\beta}{2} \nabla^2 F(\mathbf{w}_t)  \nonumber \\
 & + \frac{5\beta}{2}\underbrace{\sum_{k=1}^{K} \frac{\pi_t^k}{A} \left\| \nabla F(\mathbf{w}_t) -\nabla F(\mathbf{w}_{t-\tau_t^k}) \right\|^2}_{T_1},
\end{align}
where the second inequality is based on the fact that $\langle a, b\rangle \leq \|a\|^2+\|b\|^2$, the third inequality is based on the fact that $\langle a, b\rangle = \frac{1}{2} (\|a\|^2+\|b\|^2 - \|a-b\|^2)$, the third inequality is also derived from the Cauchy-Schwarz inequality, and the last equality is based on the fact that $\sum_{k=1}^{K} \frac{\pi_t^k}{A} = 1$. Our next step is to continuously find the upper bound of $T_1$, we have
\begin{align}
  T_1 \leq & \sum_{k\in\mathcal{K}} \frac{ \pi_t^k}{A} \|L_F(\mathbf{w}_t - \mathbf{w}_{t-\tau_t^k})\|^2 \nonumber \\
  \leq & L_F^2 \max_{k\in\mathcal{K}} \|\mathbf{w}_t - \mathbf{w}_{t-\tau_t^k}\|^2, \nonumber \\
  = & L_F^2 \|\mathbf{w}_t - \mathbf{w}_{t-\tau_t^{\xi}}\|^2,
\end{align}
where $\xi = \arg\max_{k\in\mathcal{K}} \|\mathbf{w}_t - \mathbf{w}_{t-\tau_t^k}\|^2$, the first inequality is derived from Lemma 1, that $F(\mathbf{w})$ is $L_F$-smooth, and the second inequality is derived from the fact that $\frac{1}{\sum_{k=1}^{K} \pi_t^k} \sum_{k=1}^{K} \|\pi_t^k a_k \|\leq \max_{k\in\mathcal{K}} \|a_k\|$. It follows with
\begin{align}
  T_1 = & L_F^2 \|\mathbf{w}_t - \mathbf{w}_{t-\tau_t^{\xi}}\|^2  \nonumber \\
  = & L_F^2 \left\|\sum_{j=t-\tau_t^\xi}^{t-1} (\mathbf{w}_{t+1} - \mathbf{w}_t)\right\|^2 \nonumber \\
  = & L_F^2 \left\|\sum_{j=t-\tau_t^\xi}^{t-1} \beta \sum_{k=1}^K \sum_{i=1}^{n_k} \frac{\pi_t^k}{A n_k} \nabla F_{k,i} (\mathbf{w}_{t-\tau_t^k})\right\|^2 \nonumber \\
  \leq & L_F^2\beta^2 S \sum_{j=t-S}^{t-1}\left\| \sum_{k=1}^{K} \sum_{i=1}^{n_k} \frac{\pi_t^k}{A n_k} \nabla F_{k,i} (\mathbf{w}_{t-\tau_t^k}) \right\|^2
  \nonumber \\
  = & S \sum_{j=t-S}^{t-1}\left\| \sum_{k=1}^{K} \frac{\pi_t^k}{A} \nabla F_{k} (\mathbf{w}_{t-\tau_t^k}) \right\|^2 \nonumber \\
  \leq & S \sum_{j=t-S}^{t-1} \left( \sum_{k=1}^{K} \frac{\pi_t^k}{A} \right) \left( \sum_{k=1}^{K} \frac{\pi_t^k}{A}\| \nabla F_{k} (\mathbf{w}_{t-\tau_t^k})\|^2 \right) \nonumber \\
  = & S \sum_{j=t-S}^{t-1}\sum_{k=1}^{K} \frac{\pi_t^k}{A}\| \nabla F_{k} (\mathbf{w}_{t-\tau_t^k})\|^2,
\end{align}
where the first inequality is derived from Assumption~\ref{assum:1} that $\tau_t^{\xi}$ is upper bounded by $S$ and the fact that $\|\sum_{k=1}^{K} a_k\|^2 \leq K \sum_{k=1}^{K} \|a_k\|^2$, while the second inequality is based on the Cauchy-Schwarz inequality with $a_k = \sqrt{\frac{\pi_t^k}{A}}$ and $b_k = \sqrt{\frac{\pi_t^k}{A}} \nabla F_{k} (\mathbf{w}_{t-\tau_t^k})$. It follows with
\begin{align}
  T_1 = & S \sum_{j=t-S}^{t-1}\sum_{k=1}^{K} \frac{\pi_t^k}{A}\| Y + \nabla F (\mathbf{w}_{t-\tau_t^k}) \|^2 \nonumber \\
  \leq & 2S \sum_{j=t-S}^{t-1} \sum_{k=1}^{K} \frac{\pi_t^k}{A} \|Y\|^2 \nonumber \\
  & + 2S \sum_{j=t-S}^{t-1} \sum_{k=1}^{K}\frac{\pi_t^k}{A} \nabla^2 F (\mathbf{w}_{t-\tau_t^k}) \nonumber \\
  \leq & \frac{2S^2K\gamma_F^2}{A} + 2S^2 \sum_{k=1}^{K}\frac{\pi_t^k}{A} \|\nabla F (\mathbf{w}_{t-\tau_t^k})\|^2.
\end{align}
Now getting back to (\ref{equ:convergence_rate}), we have
\begin{align}
  & F(\mathbf{w}_{t+1}) - F(\mathbf{w}_t) \nonumber \\
  \leq & \frac{10 \beta K\gamma_F^2}{A} - \frac{\beta}{2} \nabla^2 F(\mathbf{w}_t) + \frac{5\beta S^2K \gamma_F^2}{A} \nonumber \\
  & + \frac{5 \beta S^2}{A} \sum_{k=1}^{K} \pi_t^k \| \nabla F (\mathbf{w}_{t-\tau_t^k})\|^2 \nonumber \\
  \leq & \frac{10 \beta K\gamma_F^2}{A} +
  \frac{5\beta S^2K \gamma_F^2}{A} + \frac{5 \beta S^2}{A} \sum_{k=1}^{K} \pi_t^k \| \nabla F (\mathbf{w}_{t-\tau_t^k})\|^2 \nonumber \\
  = & \phi \sum_{k=1}^{K} \pi_t^k \| \nabla F (\mathbf{w}_{t-\tau_t^k})\|^2 + \nu.
\end{align}
As a result, the desired upper bound is obtained.

{
\footnotesize
\bibliographystyle{IEEEtran}
\bibliography{IEEEabrv,IEEEexample}
}

\end{document}